\documentclass[10pt,twocolumn,letterpaper]{article}

\usepackage[pagenumbers]{cvpr} 
\usepackage{multirow}
\usepackage{bbding}
\usepackage{caption}
\usepackage[dvipsnames]{xcolor}
\definecolor{cvprblue}{rgb}{0.21,0.49,0.74}
\usepackage[pagebackref,breaklinks,colorlinks,citecolor=cvprblue]{hyperref}
\usepackage{colortbl}

\definecolor{lightred}{rgb}{1, 0.5, 0.5}
\definecolor{lightblue}{rgb}{0.5, 0.6, 1}

\def\paperID{5913} 
\def\confName{CVPR}
\def\confYear{2025}

\def\method{DreamForge}

\title{DreamForge: Motion-Aware Autoregressive Video Generation \\for Multiview Driving Scenes}

\author{Jianbiao Mei$^{1,2,*}$, Tao Hu$^{2,3,*}$, Xuemeng Yang$^2$, Licheng Wen$^2$,  Yu Yang$^1$, Tiantian Wei$^{2,4}$, \\ Yukai Ma$^{1,2}$, Min Dou$^2$, Botian Shi$^{2,{\text{\Envelope}}}$, Yong Liu$^{1,{\text{\Envelope}}}$ \vspace{7pt} \\ 
$^{1}$ Zhejiang University
$^{2}$ Shanghai Artificial Intelligence Laboratory \\
$^{3}$ University of Science and Technology of China
$^{4}$ Technical University of Munich
}

\begin{document}
\maketitle

\renewcommand{\thefootnote}{\relax} 
\footnotetext{\noindent *~Equal contribution \quad \Envelope~Corresponding author}

\begin{abstract}
Recent advances in diffusion models have improved controllable streetscape generation and supported downstream perception and planning tasks. However, challenges remain in accurately modeling driving scenes and generating long videos. To alleviate these issues, we propose {\method}, an advanced diffusion-based autoregressive video generation model tailored for 3D-controllable long-term generation. To enhance the lane and foreground generation, we introduce perspective guidance and design object-wise position encoding to incorporate local 3D correlation and improve foreground object modeling. We also propose motion-aware temporal attention to capture motion cues and appearance changes in videos. By leveraging motion frames and an autoregressive generation paradigm, we can autoregressively generate long videos (over 200 frames) using a model trained in short sequences, achieving superior quality compared to the baseline in 16-frame video evaluations. Finally, we integrate our method with the realistic simulator DriveArena to provide more reliable open-loop and closed-loop evaluations for vision-based driving agents. Project Page: \small{\url{https://pjlab-adg.github.io/DriveArena/dreamforge}}.
\end{abstract}
    
\section{Introduction}
\label{intro}
With the emergence of large-scale datasets~\cite{caesar2020nuscenes, caesar2021nuplan, sun2020scalability} and growing demands for practical applications, autonomous driving (AD) algorithms have experienced remarkable advancements in recent decades. These advances have driven a shift from traditional modular pipelines~\cite{yin2021center, guo2023scenedm, li2023logonet} to end-to-end models~\cite{hu2023planning, jiang2023vad, yang2024driving}, as well as the incorporation of knowledge-driven approaches~\cite{wen2023dilu,fu2024drive,mei2024continuously}. Despite achieving impressive performance on various benchmarks, significant challenges such as generalization and handling corner cases remain, largely due to the limited data diversity in these benchmarks and the lack of a realistic simulation platform \cite{yang2024drivearena,li2024ego}.

To enhance the diversity of driving scenes and facilitate downstream perception and planning tasks, recent approaches \cite{yan2024oasim, yan2024street, gao2023magicdrive} have leveraged generative technologies such as NeRF \cite{mildenhall2021nerf}, 3D GS \cite{kerbl20233d}, and diffusion models \cite{dhariwal2021diffusion} to generate novel multiview driving scenes. Among these, diffusion-based methods \cite{wen2024panacea, gao2023magicdrive, li2023drivingdiffusion, gao2024vista, xie2024x} have attracted significant attention for their ability to generate diverse, high-fidelity scenarios with flexible control conditions. However, these methods still encounter challenges, such as modeling geometrically and contextually accurate driving scenes and maintaining temporal coherence across long videos, which may affect their effectiveness in practical applications. On the other hand, these methods primarily use their pre-trained models for data augmentation in downstream tasks, and few methods \cite{zhou2024simgen, yang2024drivearena, yan2024drivingsphere} involve the exploration of diffusion-based models for realistic generative simulations, which capture real-world visual and physical aspects, facilitate scalable scene generation, and support the ongoing development of AD algorithms within closed-loop systems.

To alleviate the above issues, following \cite{gao2023magicdrive, wen2024panacea}, we design a diffusion-based framework, named {\method} for multiview driving scene generation. Specifically, our {\method} leverages flexible control conditions, e.g., road layouts and 3D bounding boxes, along with textual inputs, to generate geometrically and contextually accurate driving scenarios, maintaining cross-view and temporal consistency. By integrating perspective guidance, object-wise position encoding, and motion-aware autoregressive generation into conditional diffusion models \cite{dhariwal2021diffusion, blattmann2023stable}, our framework achieves significant improvements in several aspects:
(1) Better controllability. We can not only control the generation of scenes with varying weather conditions and styles through texts, road layouts, and boxes but also improve street and foreground generation by perspective guidance (PG) and object-wise position encoding (OPE). The PG assists the network in learning to generate geometrically and contextually accurate driving scenes. The designed OPE enhances foreground modeling and naturally introduces local 3D correlation. (2) Better coherence. By learning motion cues from motion frame, ego pose, and feature differences and generating videos in an autoregressive manner, our {\method} can generate multiview videos with flexible lengths using a model trained with short sequences while maintaining temporal coherence. Experiments demonstrate that we can generate long videos exceeding 200 frames using only a model trained in short sequences and achieve better generation quality than the baseline in 16-frame video evaluations. In particular, our proposed {\method} can adapt to various generative base models, such as SD V1.5 \cite{blattmann2023stable} and DiT \cite{zheng2024open}, demonstrating its broader application to the autonomous driving community.

Moreover, we enhance this work by integrating our {\method} into the recent modular closed-loop generative simulation platform, DriveArena \cite{yang2024drivearena}, to explore the application of diffusion-based generative models in autonomous driving simulation. By integrating with the simulation platform, our approach offers improved scalability, which can seamlessly adapt to generating dynamic driving scenes for road networks in any city worldwide and serves as a more coherent scene render for both open-loop and closed-loop evaluations of vision-based AD algorithms.

Our contributions can be summarized as follows:

$\bullet$  We introduce perspective guidance and develop object-wise position encoding to enhance street and foreground generation. This innovative object-wise position encoding improves foreground modeling and inherently provides local 3D correlation, leading to better object generation.

$\bullet$ We propose motion-aware temporal attention to incorporate motion cues and understand the appearance changes of the video. Besides, by utilizing motion frames and an autoregressive generation paradigm, we achieve long video generation with a model trained on short sequences.

$\bullet$ We integrate the proposed {\method} with a realistic simulation platform to enhance coherent driving scene generation and offer more reliable open-loop and closed-loop evaluation for vision-based driving agents.
\section{Related Work}
\label{Rel. Work} 
\subsection{Autoregressive Video Generation}
Generating long video sequences with diffusion models is often constrained by fixed-length training due to GPU memory limitations, leading to performance degradation when extending beyond trained sequence lengths \cite{chen2023seine}. Autoregressive video generation has emerged as a promising alternative \cite{ho2022video,xing2023survey,chen2023seine}, enabling sequential prediction of future frames conditioned on prior clips \cite{gao2024vid} to produce extended, temporally coherent videos. Recent advancements in autoregressive video diffusion models have introduced various conditioning mechanisms to incorporate previous frames into the generation process, such as adaptive layer normalization \cite{voleti2022mcvd}, cross-attention \cite{henschel2024streamingt2v}, and temporal or channel-wise concatenation \cite{zeng2024make,weng2024art} in noisy latent spaces.

Recent works \cite{gao2024vista,wang2023drivedreamer} in autonomous driving also applied autoregressive video generation to forecast monocular scenarios. Unlike them, we propose a motion-aware autoregressive paradigm that learns motion cues from motion frames, ego poses, and feature differences to better understand appearance changes in long-term multiview videos.

\subsection{Autonomous Driving Scene Generation} 
For driving scene generation, some studies \cite{yan2024street, turki2023suds, lu2023urban, yan2024oasim, zhou2024drivinggaussian} use NeRF \cite{mildenhall2021nerf} and 3D GS \cite{kerbl20233d} for novel view synthesis by reconstructing scenes from logged videos, which often struggle with diverse weather and road layouts. 
On the other hand, recent advances in diffusion models \cite{dhariwal2021diffusion} have established them as leading approaches \cite{gao2024vista,li2023drivingdiffusion,yang2024generalized,wen2024panacea,huang2024subjectdrive, ma2024unleashing, wang2024driving, lu2025wovogen, li2024uniscene, jiang2024dive} for the generation of high-fidelity, diverse driving scenes through progressive denoising \cite{song2020denoising,nichol2021improved}. 
For example, several methods \cite{gao2024vista,hu2023gaia,yang2024generalized} focused on the monocular diffusion-based world model, with ego actions to control ego-vehicle behavior and generate future scenes. 
DriveDreamer \cite{wang2023drivedreamer} and MagicDrive \cite{gao2023magicdrive} employ HDmap and 3D box to enable more controllable scene generation. Recent methods, e.g., Panacea \cite{wen2024panacea}, DrivingDiffusion \cite{li2023drivingdiffusion}, and SubjectDrive \cite{huang2024subjectdrive}, further advance 3D-controllable multiview video generation. 
Unlike these methods, we integrate perspective guidance, object-wise position encoding, and motion-aware autoregressive generation into diffusion models, resulting in significant improvements in both controllability and temporal coherence for long multiview video generation.
\begin{figure*}[htbp]
    \centering
    \includegraphics[width=0.93\linewidth]{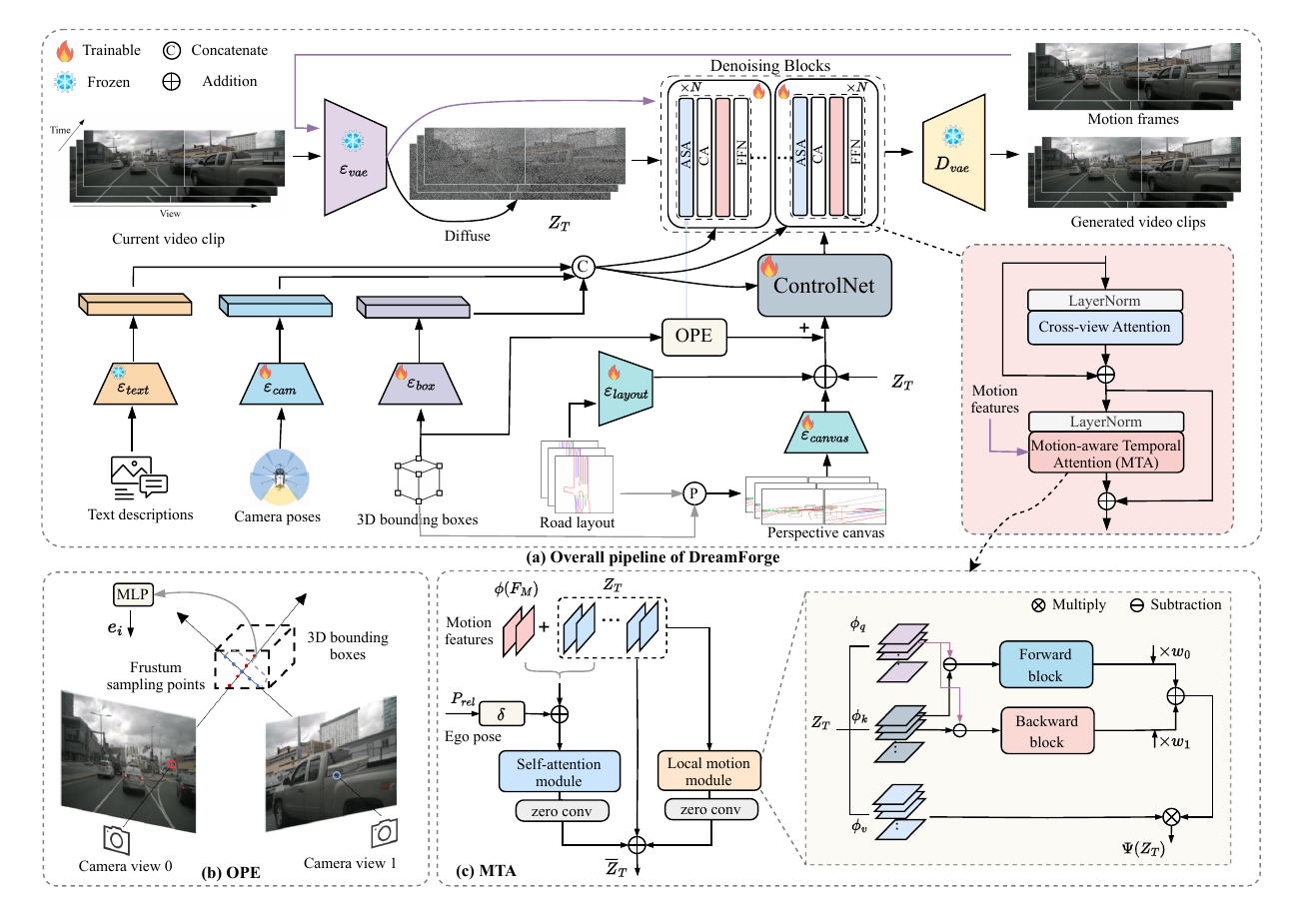}
    \vspace{-10pt}
    \caption{(a) Overall framework. During the denoising process, {\method} leverages various conditions to enhance the modeling of driving scenes. Additionally, we introduce perspective guidance and incorporate object-wise position encoding (OPE) to improve street and foreground generation. We also implement motion-aware attention (MTA) to enhance temporal coherence, supporting long-term video generation through autoregression. ``P" denotes the perspective projection. (b) The overall procedure of OPE. We only encode frustum sampling points in the 3D bounding boxes into the object position embedding. (c) The detailed architecture of MTA, which learns motion cues from motion frames, ego poses, and bidirectional feature differences.}
    \label{fig:generator}
    \vspace{-12pt}
\end{figure*}

\section{Methodology}
\label{method}
We present our proposed {\method} in Fig.~\ref{fig:generator} (a). It features diverse conditional encodings for improved controllability, perspective guidance, and object-wise position encoding for enhanced street and foreground generation (Sec.\ref{condition}). Also, a motion-aware temporal attention module and autoregressive generation are designed to enable seamless video generation (Sec.\ref{temporal}), allowing integration into a simulation platform for broader applications (Sec.\ref{simulation}).

\subsection{Improved Conditional Controllability}
\label{condition}
{\method} encodes diverse conditions for generating controllable driving videos. Additionally, we explicitly project road layouts and bounding boxes into the camera views for perspective guidance and devise the object-wise position encoding for foreground enhancement, which improves the controllability of street and foreground object generation.

\noindent\textbf{Conditional Encoding with ControlNet.} 
Similar to MagicDrive \cite{gao2023magicdrive}, we utilize scene-level descriptions, camera poses, 3D bounding boxes of foreground objects, and the road layout of background elements as various forms of conditional encoding for controllable generation. 
Specifically, for scene-level encoding, we first enrich the text descriptions using GPT-4 and then utilize the CLIP text encoder~\cite{radford2021learning} ($\mathcal{E}_{text}$) to extract the text embeddings $e_{text}$ from these descriptions. The camera poses $\{ \mathbf{K} \in \mathbb{R}^{3 \times 3}, \mathbf{R} \in \mathbb{R}^{3 \times 3}, \mathbf{T} \in \mathbb{R}^{3 \times 1} \}$ of each camera are encoded to $e_{cam}$ by Fourier Embedding~\cite{mildenhall2021nerf} and MLP ($\mathcal{E}_{cam}$), where $\mathbf{K}$, $\mathbf{R}$, $\mathbf{T}$ represent camera intrinsic, rotations and translations respectively. For 3D boxes encoding, label embeddings are first extracted from the class labels using a text encoder. Coordinate embeddings are derived from the eight vertices of the 3D box through Fourier Embedding and MLP. Finally, both label and coordinate embeddings are combined and compressed into the final box embeddings $e_{box}$ using MLP. These embeddings, $e_{text}$, $e_{cam}$, and $e_{box}$, have the same dimensions and are concatenated before being fed into the ControlNet \cite{zhang2023adding} and denoising blocks, as shown in Fig.~\ref{fig:generator} (a). As for the road layout encoding, the 2D grid-formatted road layouts are processed through a ConvNet ($\mathcal{E}_{layout}$) to produce layout embeddings $e_{layout}$, which are then combined with the noised latents and fed into the ControlNet.
\vspace{-10pt}

\begin{figure}[t]
    \centering
    \includegraphics[width=0.9\linewidth]{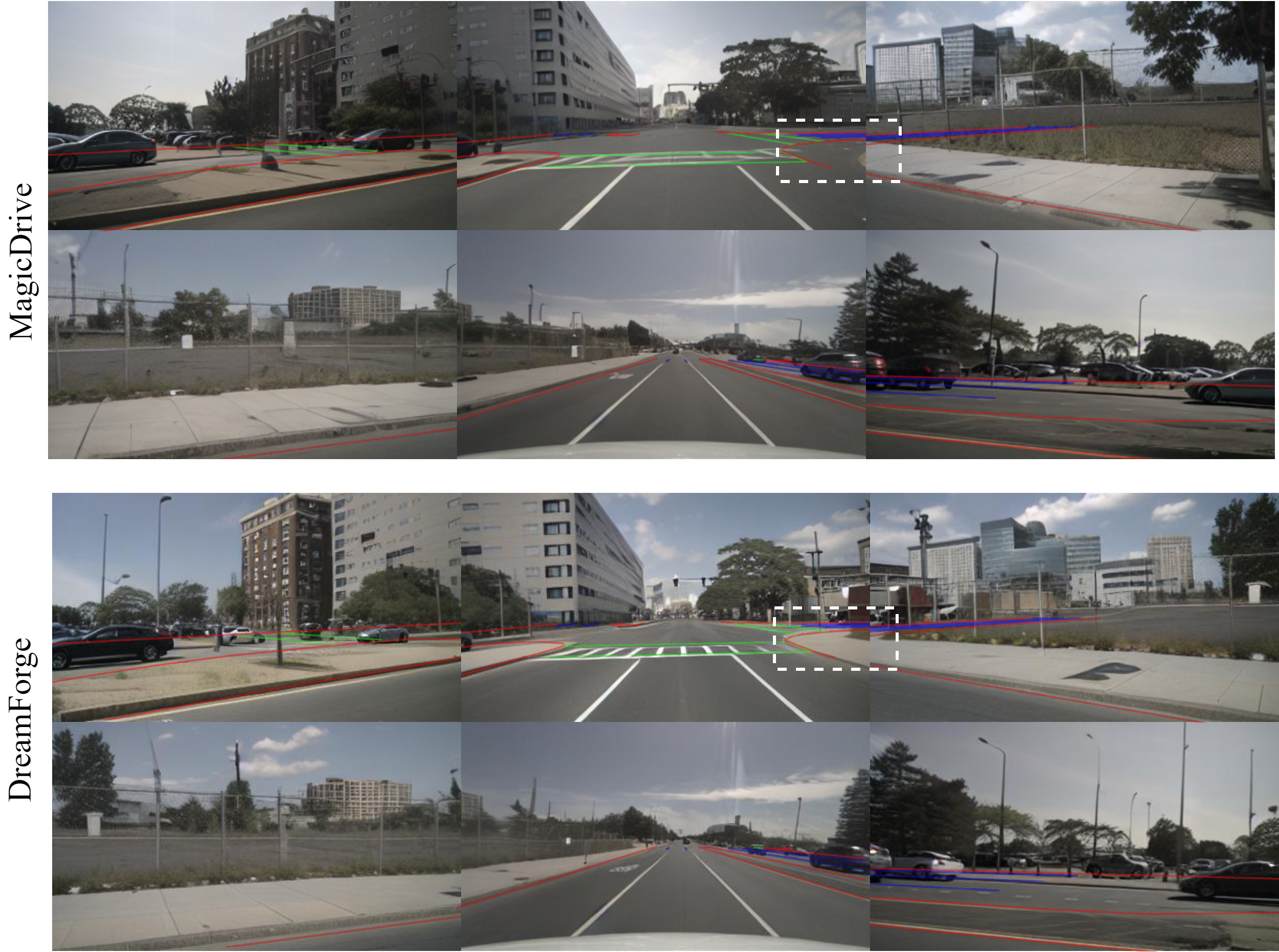}
    \caption{Visual Comparison. Our {\method} produces more geometrically accurate images due to the perspective guidance.}
    \label{fig:layout}
    \vspace{-15pt}
\end{figure}

\noindent\textbf{Perspective Guidance.} As mentioned above, ControlNet encodes rich 3D information and camera poses, which could theoretically allow it to perform view transformation implicitly \cite{gao2023magicdrive}; however, our experiments found that this implicit learning struggles to generate surround-view images that accurately align with the road layout, particularly in distant and complex areas, as illustrated in Fig.~\ref{fig:layout}. Therefore, we further project the road layout and 3D boxes into the camera view using the camera poses to explicitly provide perspective guidance for position constraints and reduce the network's difficulty in learning to generate geometrically and contextually accurate driving scenes. To this end, the contents of each category in the road layouts and the 3D boxes are projected onto the image plane of each camera to obtain the road canvas and the box canvas, respectively. Specifically, we plot the contents of each category on a dedicated channel, encoding road and box classes in one-hot maps where 1 indicates the presence of the category and 0 otherwise. This approach enforces clear and structured constraints on the perspective view, enabling the model to effectively generate diverse elements in complex scenes. Subsequently, these canvases are concatenated to create the perspective canvas, which is encoded with ConvNet ($\mathcal{E}_{canvas}$) to form the canvas embeddings $e_{canvas}$. The canvas embeddings are merged with the noised latents, and then input into ControlNet, as shown in Fig.~\ref{fig:generator} (a).

\noindent\textbf{Object-wise Position Encoding.} In addition to introducing a cross-view attention module \cite{gao2023magicdrive, wen2024panacea} that intuitively aggregates information globally from adjacent views to ensure multiview consistency, we also propose an object-wise position encoding to incorporate 3D object embeddings, enhancing foreground modeling and inherently providing local 3D correlation for improved object consistency, as shown in Fig.~\ref{fig:vis_0m} and Appendix \ref{Appendix 4.2}. 
Following \cite{liu2022petr, chen2020dsgn}, by using the transformation matrix of the camera of the $i$-th view, we transform the points $\textbf{P}_i^c\in \mathbb{R}^{W_F\times H_F \times D \times 3}$ in the discrete camera frustum space of size $(W_F, H_F, D)$ into the common 3D world space for 3D coordinates $\textbf{P}^{3d}_i$. Subsequently, $\{\textbf{P}^{3d}_i\}_{i=1}^{N_c}$ from all $N_c$ camera views are aggregated and normalized to the range $[0, 1]$ within the region of interest, producing normalized 3D points $\textbf{P}^{3d}\in \mathbb{R}^{N_c\times W_F\times H_F \times (D \times 3)}$. Then, we utilize the 3D boxes to generate the 3D masks $\textbf{M}^{3d}$, which indicate the foreground objects in points $\textbf{P}^{3d}$. Finally, the 3D points $\textbf{P}^{3d}$ and masks $\textbf{M}^{3d}$ are fed into the position encoder to obtain the 3D object position embeddings $\textbf{E}_{o} \in \mathbb{R}^{N_c \times W_F\times H_F \times C}$ of $N_c$ camera views. The position encoder is constructed with a stack of MLPs and aggregates the point features along the sampling ray. The detailed procedure can be expressed as:
\begin{equation}
\small
    \textbf{E}_{o} = \rm{MLP}(\textbf{P}^{3d}\cdot\textbf{M}^{3d})
\end{equation}

As shown in Fig.~\ref{fig:generator} (b), through using the foreground mask, only points within the 3D bounding boxes are encoded into the object position embedding. Since the 3D world space is shared among all views, the embeddings derived from different perspectives of the same object exhibit 3D correlation. Subsequently, as illustrated in Fig.~\ref{fig:generator} (a) the embeddings $\textbf{E}_{o}$ are added to the latents and fed into ControlNet to enhance the object representation and establish local object correspondence across different views.

\noindent\textbf{Augmented Spatial Attention.} We further incorporate object-wise position encoding into the self-attention module of the denoising blocks to enhance its capabilities of extracting object appearances. Specifically, given features $\textbf{Z}_s$, we add the embeddings $\textbf{E}_{o}$ to the $\textbf{Z}_s$ before feeding them into the self-attention for spatial attention. The operation of augmented spatial attention (ASA) is formulated as follows:
\begin{equation}
\small
    \textbf{Z}'_s = \rm{SelfAttn}(\textbf{Z}_s+\textbf{E}_o)
\end{equation}
Note that we reuse the self-attention layer of the denoising blocks and only fine-tune the linear layer for query mapping of the self-attention module, thereby avoiding the introduction of additional parameters and resulting in minimal increases in computational overhead during inference.

\subsection{Motion-aware Autoregressive Generation}
\label{temporal}
Recent multiview driving scene generation works \cite{gao2023magicdrive, huang2024subjectdrive} focus on fixed-length video generation but face challenges with extended videos due to memory limitations and poor temporal consistency. Some methods \cite{wen2024panacea, li2023drivingdiffusion} use keyframes as control conditions and enhance coherence through sliding windows, yet motion cues and temporal modeling remain insufficient for long video generation. We design the motion-aware temporal attention to improve consistency by integrating motion cues from historical frames, ego poses, and feature differences, enabling our method to achieve effective autoregressive video generation.
\vspace{-10pt}

\paragraph{Motion-aware Temporal Attention.}
Let $\{\textbf{I}_i\}_{i=-M}^{-1}$ represent the $M$ motion frames sampled from the previous video clip. As shown in Fig.~\ref{fig:generator} (a), these motion frames are processed by the VAE encoder to extract motion latents, which are then fed into the blocks using shared parameters with the denoising blocks to generate multi-resolution motion features. During the denoising process, these motion features are concatenated with the corresponding noised latents to compute temporal attention. Additionally, we encode the relative poses between adjacent frames into the motion embedding for ego-motion cues and propose learning bidirectional local motion within the video clip to help the model understand changes in the background. 
Fig.~\ref{fig:generator} (c) illustrates the detailed architecture of the motion-aware temporal attention module (MTA).
Specifically, given the motion features $\textbf{F}_M = \{\textbf{f}_{-M},...,\textbf{f}_{-1}\} \in \mathbb{R}^{HW\times M\times C}$ and the noised latents $\textbf{Z}_T = \{\textbf{z}_0, ..., \textbf{z}_{T-1}\} \in \mathbb{R}^{HW\times T\times C}$, where $M$, $T$, $H$, $W$, and $C$ denote the motion length, video length, spatial height, width, and number of channels, respectively, we calculate the MTA as: 
\begin{equation}
\small
    \textbf{Z}_{MT} = [\phi(\textbf{F}_M), \textbf{Z}_T]
\end{equation}
\begin{equation}
\small
    \overline{\textbf{Z}}_{MT} = \textbf{Z}_{MT} + {\rm ZeroConv}({\rm SelfAttn}(\textbf{Z}_{MT} + \delta(\textbf{P}_{rel})))
\end{equation}
\begin{equation}
\small
    \overline{\textbf{Z}}_{T} = \overline{\textbf{Z}}_{MT}[M\colon] + {\rm ZeroConv}({\Psi}(\textbf{Z}_{T}))
\end{equation} where $\phi$ is a linear adapter, $\delta$ denotes the MLP used for ego motion encoding, and $\textbf{P}_{rel}$ represents the relative poses between adjacent frames. Note that the relative pose is set to the identity matrix for the initial motion frame. $\Psi$ is the bidirectional local motion module to aggregate motion cues from the forward and backward feature differences of the video clip. In the forward process, we subtract the previous frame's features from the current frame and then use convolutions to automatically learn information from the adjacent feature differences, and vice versa. Given the $t$-th frame $\textbf{z}_t$ from $\textbf{Z}_T$, the detailed procedure of $\Psi$ can be expressed as:
\begin{equation}
\small
    \textbf{z}_{d_0, t} = \phi_q(\textbf{z}_t) - \phi_k(\textbf{z}_{t-1}); \quad \textbf{z}_{d_1, t} = \phi_q(\textbf{z}_{t}) - \phi_k(\textbf{z}_{t+1})
\end{equation}
\begin{equation}
\small
    \Psi(\textbf{z}_t) = [w_0\cdot \gamma_f(\textbf{z}_{d_0, t}) + w_1\cdot \gamma_b(\textbf{z}_{d_1, t})]\cdot \phi_v(\textbf{z}_{t})
\end{equation} where $\phi_*$ is the linear projection layer, $\gamma_f$ and $\gamma_b$ represent the forward and backward blocks, respectively, each consisting of two-layer convolutions. The parameters $w_0$ and $w_1$ are learnable. In the first frame, we set the forward feature differences to zero, while in the last frame, we set the backward feature differences to zero.
\vspace{-10pt}

\paragraph{Autoregressive Video Generation.} \label{autoregressive}
To facilitate online inference and streaming video generation while maintaining temporal coherence, we employ an autoregressive generation pipeline. During inference, we sample previously generated images as motion frames and calculate the corresponding relative ego poses to provide motion cues. This method enables the diffusion model to generate the current video clip with enhanced consistency, ensuring smoother transitions and improved coherence with previously generated frames. By utilizing motion frames, our method eliminates the need for a sliding window, thereby avoiding redundant generation. 
Our method also supports an optional post-processing strategy using sliding windows to further enhance temporal coherence between adjacent video clips. Please see the Appendix \ref{Appendix 2.1} for more details. 

\subsection{World Dreamer for Closed-loop Simulation}
\label{simulation}
Leveraging the enhanced controllability and consistency, we integrate our {\method} into a closed-loop simulation platform, DriveArena \cite{yang2024drivearena}, to investigate the application of diffusion-based generative models in driving simulations.

\begin{figure}[t]
    \centering
    \includegraphics[width=0.92\linewidth]{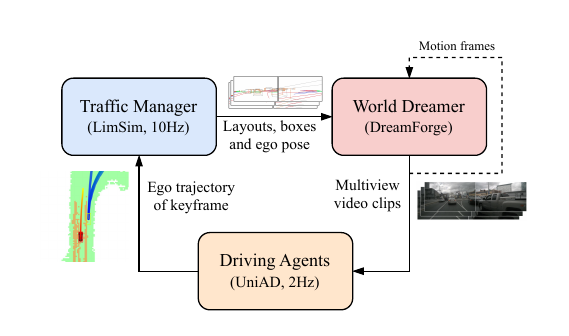}
    \caption{The closed-loop simulation platform DriveArena \cite{yang2024drivearena} utilizes LimSim \cite{wenl2023limsim} to parse HD maps, manage traffic flow, detect collisions, and generate road layouts, vehicle boxes, and ego poses for driving scene generation. We upgrade the Wolrd Dreamer with our {\method} for better temporal coherence.}
    \label{fig:arena}
    \vspace{-15pt}
\end{figure}

\noindent\textbf{Closed-loop Simulation Workflow.}
DriveArena offers a modular platform that can be integrated with different world dreamer and vision-based driving agents for both open-loop and closed-loop simulations.
As shown in Fig.~\ref{fig:arena}, in each loop: (1) The Traffic Manager receives the ego trajectory output from the driving agent (in closed-loop mode) or generates it itself (in open-loop mode), managing the movements of all vehicles and creating scene layouts. (2) World Dreamer (i.e., {\method}) utilizes the received road layouts and vehicle boxes as control conditions to generate surround-view images. (3) The Driving Agent takes the visual images as input and directly plans the ego trajectory, which is then sent to the Traffic Manager for the next rollout.
With features that enhance both controllability and long-term temporal consistency in the generated images, our {\method} serves as an effective world image renderer for autonomous driving simulations.

\section{Experiments}
\label{exp}
Our {\method} is built on SD V1.5 \cite{rombach2022high} by default. We also offer a version using DiT \cite{zheng2024open} and 3D VAE \cite{yang2024cogvideox} as the base model, as detailed in Sec.\ref{exp:ab} and Table \ref{tab:abresult}. The default input resolution for each view is $224 \times 400$, with a video length $T$ of 7 and motion frames length $M$ of 2. Please see more details about implementation, datasets, and metrics in the Appendix; and find more demos on project page.

\subsection{Quantitative Comparison}
\paragraph{Fidelity Validation on the nuScenes Dataset.} We apply the proposed {\method} to generate realistic multiview scenes using annotations from the nuScene validation set. We offer variants at different resolutions to facilitate a comprehensive comparison with recent methods. Following previous works \cite{gao2023magicdrive, swerdlow2024street}, we utilize BEVFusion \cite{liang2022bevfusion} for 3D object detection and CVT \cite{zhou2022cross} for BEV segmentation. The results are shown in Table \ref{tab:single_val} and illustrate that our method has a lower FID and better performance on downstream perception tasks at $224\times400$ resolution. For example, our {\method} exceeds the baseline by 6.74, 1.14, and 1.26 points in terms of road mIoU, vehicle mIoU, and object mAP. Moreover, by improving the input resolution, the performance of 3D object detection and BEV segmentation is further enhanced. Notably, for BEV segmentation, the performance on the generated scenes using our {\method} at $336 \times 600$ closely matches the performance on the real dataset. We also observed that the FID value increases at higher resolutions. We attribute this to the fact that higher resolution aids in detecting more objects, including small and distant ones; however, it may also complicate the generation of background details.

\begin{table}[t]
    \setlength{\tabcolsep}{2pt}
    \small
    \centering
    \resizebox{\linewidth}{!}{
    \begin{tabular}{l|c|c|cc|cc}
    \toprule
    \multirow{2}[3]{*}{\textbf{Data Source}} & \multirow{2}[3]{*}{\begin{tabular}[c]{@{}c@{}}\textbf{Synthesis}\\ \textbf{resolution}\end{tabular}} & \multirow{2}[3]{*}{\textbf{FID}$\downarrow$} & \multicolumn{2}{c|}{\textbf{BEV segmentation}} & \multicolumn{2}{c}{\textbf{3D object detection}} \\ \cmidrule{4-7} &&& Road mIoU $\uparrow$ & Vehicle mIoU $\uparrow$ & mAP $\uparrow$ & NDS $\uparrow$ \\
    \midrule
    Ori nuScenes & - & - & 73.67 & 34.82 & 35.54 & 41.21 \\
    \midrule
    DriveDreamer \cite{wang2023drivedreamer} & - & 26.80 & - & - & - & -     \\
    Panacea \cite{wen2024panacea} & 256$\times$512 & 16.96 & - & - & - & -     \\
    BEVGen \cite{swerdlow2024street} & 224$\times$400 & 25.54 & 50.20 & 5.89 & - & -     \\
    BEVControl \cite{yang2023bevcontrol} & - & 24.85 & 60.80 & 26.80 & - & -     \\ 
    MagicDrive \cite{gao2023magicdrive} & 224$\times$400 & {16.20} & 61.05 & 27.01 & 12.30 & 23.32 \\
    MagicDrive* & 224$\times$400 & 19.06 & 58.53 & 27.22 & 11.75 & 22.79 \\
    X-Drive \cite{xie2024x} & 224$\times$400 & \underline{16.01} & - & - & - & - \\
    \midrule
    \method & 224$\times$400 & \textbf{14.61} & 65.27 & 28.36 & 13.01 & 22.16 \\
    \method & 336$\times$600 & 28.77 & \underline{69.43} & \underline{32.12} & \underline{19.29} & \underline{28.88} \\
    \method & 448$\times$800 & 30.06 & \textbf{69.76} & \textbf{33.49} & \textbf{24.13} & \textbf{33.00} \\
    \bottomrule
    \end{tabular}
    }
    \caption{Comparison of generation fidelity with driving generation methods on nuScenes validation. 
    \textbf{Bold} represents the best results.  \underline{Underline} indicates the second best results. * Results are computed using the official weights.}
    \label{tab:single_val}
\end{table}

\begin{table}[t]
    \setlength{\tabcolsep}{7pt}
    \small
    \centering
    \resizebox{\linewidth}{!}{
    \begin{tabular}{l|c|c|ccc} 
    \toprule
    \textbf{Data Source} & \textbf{Resolution} & \textbf{Frames} & \textbf{FVD} $\downarrow$ & \textbf{mAP} $\uparrow$ & \textbf{mIoU} $\uparrow$ \\ 
    \cmidrule{1-6}
    Ori nuScenes & 224$\times$400  & - & - & 29.69 & 36.70  \\
    \midrule
    MagicDrive & 224$\times$400 & 16/16 & 218.1 & 11.86 & 18.34  \\
    {\method} & 224$\times$400 & 7/16 & 209.9 & 14.37 & 29.07 \\
    {\method} & 336$\times$600 & 7/16 & \textbf{197.9} & 20.03 & 31.96 \\
    {\method} & 448$\times$800 & 7/16 & 233.2 & \textbf{22.52} & \textbf{32.98} \\
    \bottomrule
    \end{tabular}}
    \caption{Comparison of generation fidelity on generated 16-frame clips from nuScene validation (tested by BEVFormer \cite{li2022bevformer}). 
     ``7/16" indicates training using 7 frames while inference with 16 frames. }
    \label{tab:vmetrics}
\end{table}

\begin{table}[t]
    \setlength{\tabcolsep}{7pt}
    \small
    \centering
    \resizebox{\linewidth}{!}{
    \begin{tabular}{l|c|c|ccc} 
    \toprule
    \textbf{Data Source} & \textbf{Resolution} & \textbf{Base Model} & \textbf{FVD} $\downarrow$ & \textbf{mAP} $\uparrow$ & \textbf{mIoU} $\uparrow$ \\ 
    \cmidrule{1-6}
    Ori nuScenes & 224$\times$400  & - & - & 29.69 & 36.70  \\
    \midrule
    MagicDrive~\cite{gao2023magicdrive} & 224$\times$400 & SD V1.5 & 218.12 & 11.86 & 18.34  \\
    MagicDrive3D~\cite{gao2024magicdrive3d} & 224$\times$400 & SD V1.5 & 210.40 & 12.05 & 18.24  \\
    {\method} & 224$\times$400 & SD V1.5 & 209.90 & 14.37 & 29.07 \\
    {\method} & 448$\times$800 & SD V1.5 & 233.20 & \textbf{22.52} & \underline{32.98} \\
    \midrule
    MagicDriveDiT~\cite{gao2024magicdrivedit} & 848$\times$1600 & 3DVAE, DiT & \textbf{94.84} & 18.17 & 20.40 \\
    {\method}$^{\dagger}$ & 448$\times$800 & 3DVAE, DiT & \underline{103.61} & \underline{19.17} & \textbf{34.36} \\
    \bottomrule
    \end{tabular}}
    \caption{Comparison of the model under different base model configurations. Metrics are computed for 16-frame video clips.}
    \label{tab:abresult} 
\end{table}

\begin{table}[tbp]
    \setlength{\tabcolsep}{8pt}
    \small
    \centering
    \resizebox{\linewidth}{!}{
    \begin{tabular}{l|lccc} 
    \toprule
    \textbf{Data Type} & \textbf{Data Source} & \textbf{mAP} $\uparrow$ & \textbf{NDS} $\uparrow$ & \textbf{mAoE} $\downarrow$ \\ 
    \cmidrule{1-5}
    Real & Ori nuScenes  & 34.5 & 46.9 & 59.4 \\
    \cmidrule{1-5}
    \multirow{2}{*}{Generated} & Panacea \cite{wen2024panacea} & 22.5 & 36.1 & 72.7 \\
    & {\method} & \textbf{26.0} & \textbf{41.1} & \textbf{62.2} \\
    \cmidrule{1-5}
    \multirow{4}{*}{Real + Generated} & DriveDreamer \cite{wang2023drivedreamer}& 35.8 & 39.5 & -  \\
     & MagicDrive \cite{gao2023magicdrive} & 35.4 & 39.8 & -  \\
     & Panacea \cite{wen2024panacea} & \textbf{37.1} & \textbf{49.2} & 54.2  \\
     & {\method} & \underline{36.6} & \underline{49.1} & \textbf{52.9}  \\
    \bottomrule
    \end{tabular}}
    \caption{Comparison of performance for the 3D object detection task (tested by StreamPETR \cite{wang2023exploring}) against other methods.}
    \label{tab:streampetr}
    \vspace{-10pt}
\end{table}

\begin{figure}[t]
    \centering
    \includegraphics[width=0.98\linewidth]{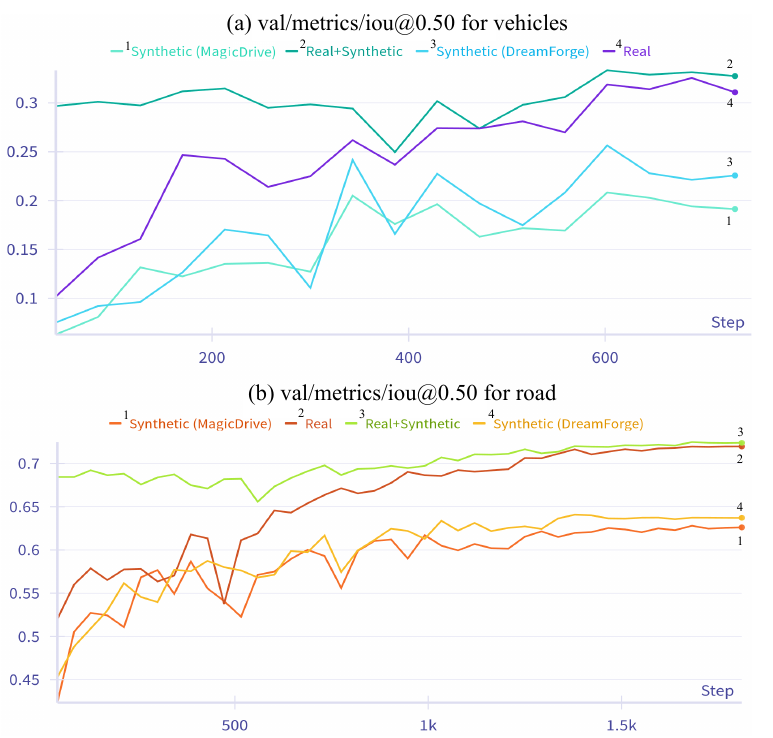}
    \caption{Validation results of map-view segmentation for vehicles (a) and road (b) during the training procedure of CVT \cite{zhou2022cross}.}
    \label{fig:cvt_train}
    \vspace{-18pt}
\end{figure}

We present a video generation fidelity comparison in Table \ref{tab:vmetrics}, evaluating the quality of the generated videos with a length of 16 frames. Compared to the baseline, i.e., temporal MagicDrive trained with 16-frame clips, our {\method} achieves a lower FVD and significantly outperforms the baseline in terms of object mAP ($+3.51$ points) and map mIoU ($+10.73$ points). Furthermore, when upgrading the resolution to $336 \times 600$, the object mAP shows a substantial increase, exceeding the baseline by 8.17 mAP. When the resolution continues to increase to $448 \times 800$, the performance of the perception results improves further, but the FVD rises. This indicates that, while fixing the training iterations, a larger resolution is more beneficial for perception models, but it also complicates the similarity in synthetic data. Notably, we generate the required clips through motion-aware autoregressive generation using only our model trained in short sequences (e.g. 7 frames). There is no need to retrain the model to produce longer videos, making our approach more efficient and resource-friendly.

\begin{table*}[tbp]
    \centering
    \small
    \setlength{\tabcolsep}{7pt}
    \resizebox{0.9\textwidth}{!}{
    \begin{tabular}{c|c|cccccc}
    \toprule
    \textbf{Scenario} & \textbf{Driving Agent} & \textbf{NC} $\uparrow$ & \textbf{DAC} $\uparrow$ & \textbf{EP} $\uparrow$ & \textbf{TTC} $\uparrow$ & \textbf{C} $\uparrow$ & \textbf{PDMS} $\uparrow$\\ \midrule
    \multirow{2}{*}{{DriveArena \cite{yang2024drivearena}}} & VAD \cite{jiang2023vad} & {0.807±0.11} & {0.950±0.05} & {0.795±0.13} & {0.800±0.12} & {0.913±0.09} & {0.683±0.12} \\
     & UniAD \cite{hu2023planning} & 0.792±0.11 & 0.942±0.04 & 0.738±0.11 & 0.771±0.12  & 0.749±0.16 & 0.636±0.08 \\
    \midrule
    \multirow{2}{*}{DriveArena* (w/ \method)} & VAD \cite{jiang2023vad} & 0.829±0.08 & 0.954±0.05 & 0.767±0.07 & 0.815±0.11 & 0.920±0.10 & 0.687±0.05 \\
     & UniAD \cite{hu2023planning} & 0.843±0.04 & 0.958±0.05 & 0.728±0.06 & 0.829±0.05 &  0.704±0.14& 0.669±0.02 \\
     \bottomrule
    \end{tabular}
    }
    \caption{Performance of driving agents in DriveArena's open-loop mode. Scenarios: \textit{1)} DriveArena's open-loop simulation sequences; \textit{2)} Open-loop simulation sequences generated by replacing the world dreamer of DriveArena with our \method. Metrics include no collisions (NC), drivable area compliance (DAC), ego progress (EP), time-to-collision (TTC), comfort (C), and PDM Score (PDMS).}
    \vspace{-10pt}
\label{tab:open-loop}
\end{table*}

\begin{table}[t]
    \centering
    \small
    \setlength{\tabcolsep}{12pt}
    \resizebox{0.98\linewidth}{!}{%
    \begin{tabular}{c|c|cc}
    \toprule
    \textbf{Route} & \textbf{Driving Agent} & \textbf{PDMS} $\uparrow$& \textbf{ADS} $\uparrow$\\ 
    \midrule
    \multirow{4}{*}{\parbox{2cm}{\texttt{singapore-}\\\texttt{onenorth}}} & VAD & 0.5315 & 0.0248 \\
    & VAD* & 0.5684 (+0.0369) & 0.0315 \\  
    & UniAD  & {0.7615} & {0.1282} \\ 
    & UniAD*  & 0.8102 (+0.0487) & 0.1197 \\    
    \midrule
    \multirow{4}{*}{\parbox{2cm}{\texttt{boston-}\\\texttt{seaport}}} & VAD & {0.5830} & 0.0352 \\
    & VAD* & 0.6140 (+0.0310) & 0.0532 \\
    & UniAD & 0.4952 & {0.0450} \\
    & UniAD* & 0.7401 (+0.2449) & 0.0760 \\
    \bottomrule
    \end{tabular}}
    \caption{Evaluation of Driving Agents' performance in closed-loop mode of DriveArena. Metrics: PDM Score (PDMS) and Arena Driving Score (ADS). * denotes replacing the world dreamer in DriveArena with our \method.}
    \label{tab:closed-loop}
    \vspace{-4pt}
\end{table}

\begin{table}[ht]
    \setlength{\tabcolsep}{2pt}
    \small
    \centering
    \resizebox{0.98\linewidth}{!}{%
    \begin{tabular}{l|c|cc|ccc|c} 
    \toprule
    \multirow{2}{*}{\textbf{Data Source}} & \multirow{2}{*}{\textbf{FID} $\downarrow$} & \multirow{2}{*}{\textbf{mAP} $\uparrow$} & \multirow{2}{*}{\textbf{NDS} $\uparrow$} & \multicolumn{4}{c}{\textbf{mIoU} $\uparrow$} \\ 
    \cmidrule{5-8}&&&& {Divider} &{ Pred crossing} & {Boundary} & Mean \\
    \cmidrule{1-8}
    Ori nuScenes & - & 34.89 & 46.99 & 43.56 & 30.93 & 43.82 & 39.44   \\ \hline
    Baseline  & 19.05 & 15.15 & 29.37 & 24.48 & 7.79 & 22.92 & 18.40  \\
    + PG & 16.03 & 16.57  & 29.50 & 33.03 & \textbf{20.99} & 36.62 & \textbf{30.21} \\
    + PG, OPE & 15.44 & 17.20 & 29.84  & 32.93 & 19.74 & \textbf{36.89} & 29.85 \\
    + PG, OPE, ASA & \textbf{14.61} & \textbf{18.37}  & \textbf{31.28} & \textbf{33.52} & 19.74 & 36.61 & 29.96 \\
    \bottomrule
    \end{tabular}}
    \caption{Comparison of generation fidelity on generate images from nuScenes validation (tested by BEVFormer \cite{li2022bevformer}). ``PG", ``OPE" and ``ASA" denote perspective guidance, object-wise position encoding, and augmented spatial attention, respectively.  
    }
    \label{tab:ab1}
    \vspace{-4pt}
\end{table}

\begin{table}[ht]
    \setlength{\tabcolsep}{3pt}
    \small
    \centering
    \resizebox{0.99\linewidth}{!}{%
    \begin{tabular}{l|c|ccc|c} 
    \toprule
    \multirow{2}{*}{\textbf{Data Source}} & \multirow{2}{*}{\textbf{mAP} $\uparrow$} & \multicolumn{4}{c}{\textbf{mIoU} $\uparrow$} \\ 
    \cmidrule{3-6}&& {Divider} &{ Pred crossing} & {Boundary} & Mean \\
    \cmidrule{1-6}
    Ori nuScenes & 29.69 & 41.39 & 28.44 & 40.25 & 36.70   \\ \hline
    Full model & 14.37 & 34.28 & 18.84 & 34.11  & 29.07  \\
    w/o LMM & 14.53  & 31.25 & 19.07 & 31.61 & 27.31 (-1.76) \\
    \bottomrule
    \end{tabular}}
    \caption{Comparison of generation fidelity on generated 16-frame video clips from nuScenes validation (tested by BEVFormer \cite{li2022bevformer}). 
    }
    \label{tab:ab2}
    \vspace{-18pt}
\end{table}

\noindent\textbf{Data Augmentation for Perception Models.} We further explore the effectiveness of our proposed method in data augmentation.  We utilize {\method} to generate additional synthetic data using annotations from the nuScenes training set to train StreamPETR \cite{wang2023exploring} and CVT \cite{zhou2022cross} for 3D object detection and map segmentation. We then evaluate their performance on the real nuScenes validation set. Detailed results are presented in Table \ref{tab:streampetr} and Fig.~\ref{fig:cvt_train}. With only generated data for training, our {\method} outperforms recent methods in both detection and segmentation metrics. Furthermore, when using synthetic data for pre-training followed by fine-tuning on real data, our method achieves performance comparable to Panacea \cite{wen2024panacea} in the detection task and significantly outperforms the baseline MagicDrive \cite{gao2023magicdrive} in both detection and segmentation tasks. Additionally, the validation curves in Fig.~\ref{fig:cvt_train} show that pre-training on synthetic data enables faster convergence during fine-tuning, leading to rapid achievement of high results.

\noindent\textbf{Open-loop and Closed-loop Evaluations.} We integrate {\method} with DriveArena \cite{yang2024drivearena} to perform open-loop and closed-loop evaluations, assessing the effect on the performance of driving agents such as UniAD \cite{hu2023planning} and VAD \cite{jiang2023vad}. 
First, we evaluate the performance of driving agents in the open-loop mode, where DriveArena simulates four routes, selecting two paths in Boston and two in Singapore. The simulation duration is 120 seconds, and the results from these four routes are utilized to calculate the mean value and standard error. The results are presented in Table \ref{tab:open-loop} and show some interesting findings: (1) For UniAD, temporal coherence brought by our DreamForge leads to a significant boost to performance, especially on the metrics of no collisions (NC) and time-to-collision (TTC). (2) In the open-loop mode of DriveArena, VAD consistently outperforms UniAD by a large margin in terms of comfort (C) in both versions of DriveArena. (3) Temporal coherence enhances the stability of performance across different routes, resulting in consistently smaller standard errors on most metrics.

\begin{figure}[t]
    \centering
    \includegraphics[width=\linewidth]{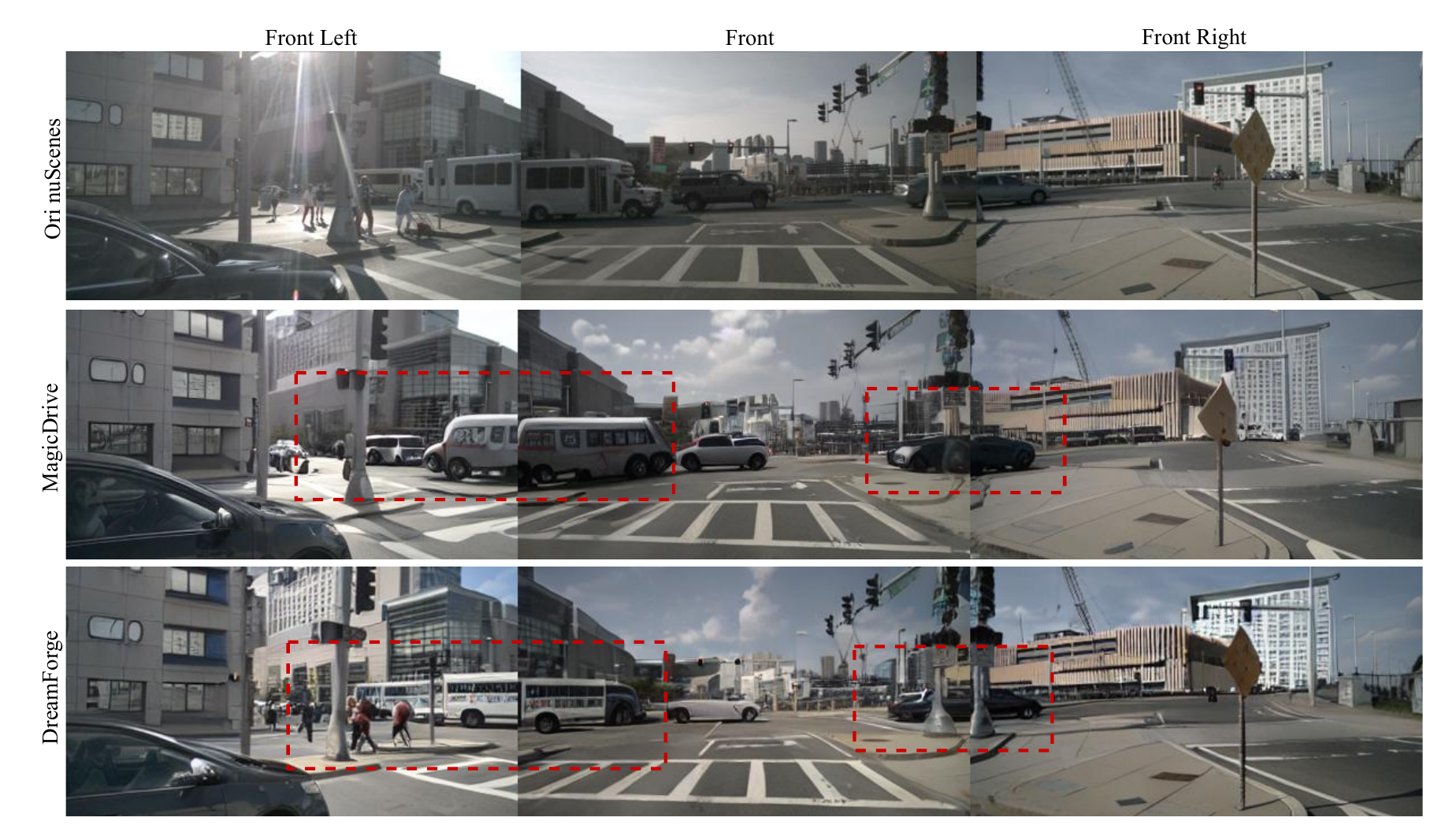}
    \caption{Visual comparison of foreground generation. The illustrations demonstrate that our {\method} achieves better foreground object generation. Please see the Appendix for more cases.}
    \label{fig:vis_0m}
    \vspace{-17pt}
\end{figure}

We further conduct experiments in DriveArena's closed-loop mode. In this mode, the trajectory outputted by the driving agents at 2 Hz, consisting of six path points over the next 3 seconds, is interpolated to create a 10 Hz trajectory, which is then directly used for ego vehicle control. Without loss of generality, we perform closed-loop testing on two representative routes in Singapore-oneorth and Boston-seaport. PDM Score (PDMS)
and Arena Drive Score (ADS) are evaluated, with detailed results presented in Table~\ref{tab:closed-loop}. From these results, we can also conclude that: (1) Upgrading DriveArena with our temporal version {\method} boosts the PDMS scores for both the routes and the driving agents. (2) UniAD consistently outperforms VAD on these routes in the closed-loop mode of the upgraded DriveArena. (3) The performance of UniAD is more susceptible to temporal coherence, which aligns with the observations made in the open-loop mode. These findings demonstrate the temporal coherence brought by our DreamForge can facilitate the application of DriveArena for realistic simulations. 

\subsection{Qualitative Results}
\paragraph{Controllable Generation.} We compare our method with the MagicDrive baseline \cite{gao2023magicdrive} to assess qualitative results. Fig.~\ref{fig:layout} shows that our method produces more geometrically accurate multiview images due to perspective guidance. Additionally, Fig.~\ref{fig:vis_0m} illustrates that our {\method} performs better in accurate object generation and maintaining consistency in street views, highlighting the effectiveness of the proposed object-wise position encoding. 
In addition, our method supports additional control conditions, such as road layouts, 3D boxes, and text prompts for generating diverse scenes, including smooth weather transitions. We also observe that ego pose influences the generated background, demonstrating the effectiveness of motion cues from ego poses. Please see Appendix \ref{Appendix 4.2} for more details.

\begin{figure}[t]
    \centering
    \includegraphics[width=\linewidth]{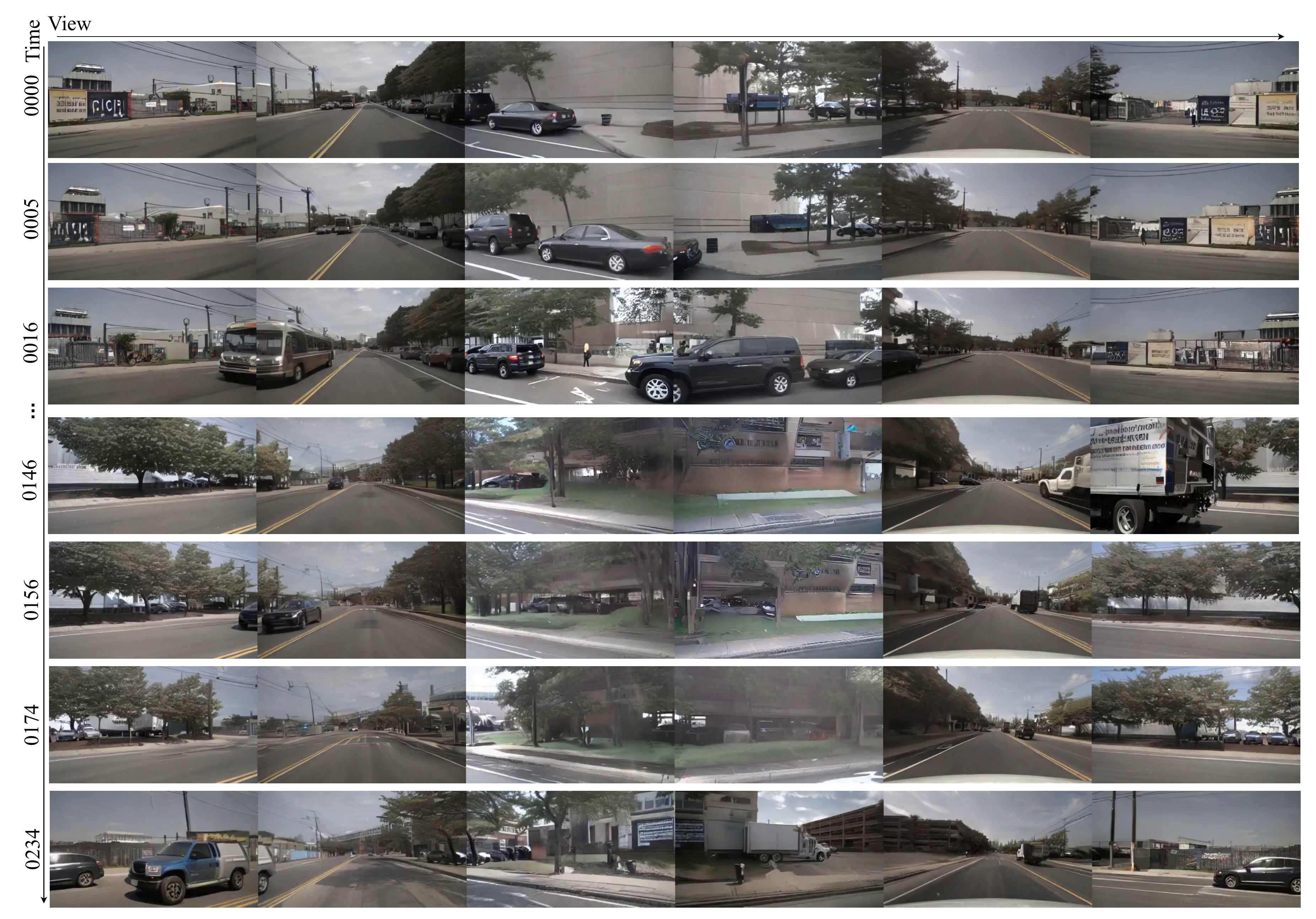}
    \caption{Long video generation. We illustrate the sampled frames from the generated long videos. }
    \label{fig:vis_1}
    \vspace{-12pt}
\end{figure}

\noindent\textbf{Long Multiview Videos.} 
Our motion-aware autoregressive generation pipeline enables the synthesis of long multiview videos (over 200 frames) using a model trained on short sequences. As shown in Fig.~\ref{fig:vis_1}, our model, conditioned on road layouts and 3D bounding boxes, produces videos at 12 Hz with high 3D controllability, fidelity, and frame consistency. This capability highlights its potential for realistic autonomous driving simulations. Please see Appendix \ref{Appendix 4.2} and the project page for more long video demo.

\subsection{Ablation Study}
We use MagicDrive \cite{gao2023magicdrive} as our baseline to evaluate performance and demonstrate the effectiveness of the fundamental components in our {\method}. Additionally, we utilize the official BEVFormer \cite{li2022bevformer} to compute the 3D object detection and map segmentation metrics.
\label{exp:ab}

\noindent\textbf{Effectiveness of Different Base Model.} To demonstrate that our proposed method can adapt to different base models, we conduct a detailed ablation study, as shown in Table \ref{tab:abresult}. We examine two configurations of the base model: SD V1.5 \cite{rombach2022high} and DiT \cite{zheng2024open, chen2023pixart} with the 3D VAE \cite{yang2024cogvideox}. The results indicate that our model consistently improves both mAP and mIoU across both configurations. Furthermore, using autoregressive generation achieves comparable FVD to the baseline when employing the same base model. Notably, employing DiT with the 3D VAE significantly enhances FVD, demonstrating the effectiveness of the temporal connections in the 3D VAE and the stronger temporal modeling capabilities of DiT. However, we also observe that using DiT with the 3D VAE results in a decline in mAP performance for object detection. We attribute this to the temporal downsampling applied to bounding boxes, which may introduce positional ambiguity, especially for dynamic objects, thereby affecting fine-grained controllability. This will be a focus for future exploration.

\noindent\textbf{Effectiveness of Perspective Guidance.} 
We conduct experiments in Table \ref{tab:ab1} and reveal that projecting road layouts and 3D bounding boxes onto the camera view for perspective guidance enhances performance across all metrics, including FID, accuracy in 3D object detection, and map segmentation. We have observed that perspective guidance significantly improves the quality of map segmentation (a 64.2\% improvement) in terms of mean value, demonstrating its effectiveness in reducing the difficulty for the network in learning to generate geometrically and contextually accurate driving scenes.

\noindent\textbf{Effectiveness of Object-wise Position Encoding.} We further validate the impact of the OPE module. The detailed ablation results are presented in Table \ref{tab:ab1}. When the object position embeddings are directly fed into ControlNet, the FID values decrease by 0.59, while the object mAP and NDS are boosted by 0.63 and 0.34 points, respectively. Further incorporating object position embeddings into the self-attention module of the denoising blocks brings an improvement of 1.17 mAP and 1.44 NDS. The FID values are also decreased by 0.83. The gains highlight the enhanced foreground generation capability of the proposed OPE module, which introduces only a slight increase in parameters due to the reuse of the self-attention layer. 
Please refer to Appendix \ref{Appendix 4.2} for more visual comparisons.

\noindent\textbf{Effectiveness of Motion-aware Temporal Attention.} 
In this section, we analyze the impact of motion-aware temporal attention (MTA). As shown in Table \ref{tab:vmetrics}, our method, leveraging MTA, achieves lower FVD and higher object mAP and segmentation mIoU compared to the baseline trained on 16-frame clips, even when generating 16-frame videos from a 7-frame setup. This highlights MTA’s effectiveness in capturing temporal dynamics. Additionally, an ablation study on the local motion module (LMM) in Table \ref{tab:ab2} reveals its ability to improve map segmentation by 1.76 points, particularly for the divider and boundary segmentation (gains of 3.03 and 2.50 points, respectively), while preserving object mAP. These results suggest that LMM effectively leverages local motion to enhance structural understanding.
\vspace{-4pt}

\section{Conclusion}
\label{conclusion}
This paper introduces {\method}, an advanced diffusion-based autoregressive model for long-term 3D-controllable video generation. By incorporating perspective guidance and object-wise position encoding, we enhance the quality of street and foreground object generation. Additionally, our motion-aware temporal attention effectively captures motion cues, enabling the generation of long videos (over 200 frames) with a model trained on short sequences, outperforming the baseline in quality. Finally, we integrate our method with DriveArena to improve simulation and provide reliable evaluations for vision-based driving agents.
{
    \small
    \bibliographystyle{ieeetr}
    \bibliography{main}
}

\newpage
\clearpage
\appendix
\section{Appendix}

\label{appendix}

\begin{table*}[ht]
\begin{center}
\resizebox{0.98\textwidth}{!}{
\small
\centering
\begin{tabular}{l|c|cc|cccc|cccc|cccc} 
\toprule
\multirow{2}{*}{\bf Data Source} & \multirow{2}[3]{*}{\begin{tabular}[c]{@{}c@{}}\textbf{Image}\\ \textbf{resolution}\end{tabular}} & \multicolumn{2}{c|}{\bf 3DOD} & \multicolumn{4}{c|}{\bf BEV Segmentation mIoU (\%)} & \multicolumn{4}{c|}{\bf L2 (m) $\downarrow$}& \multicolumn{4}{c}{\bf Col. Rate (\%) $\downarrow$} \\
\cmidrule{3-16}
                                     && mAP $\uparrow$  & NDS $\uparrow$ & Lanes $\uparrow$ & Drivable $\uparrow$ & Divider $\uparrow$ & Crossing $\uparrow$ & 1.0s & 2.0s & 3.0s & Avg.  & 1.0s   & 2.0s   & 3.0s   & Avg.     \\ 
\cmidrule{1-16}
Ori nuScenes & $896\times1600$ & 37.98 & 49.85 & 31.31 & 69.14  & 25.93 & 14.36 & 0.51 & 0.98 & 1.65 & 1.05  & 0.10 & 0.15 & 0.61 & 0.29  \\
Ori nuScenes & $224\times400$ & 31.20 & 45.22 & 29.19 & 65.83  & 23.51 & 12.99 & 0.60 & 1.10 & 1.85 & 1.18  & 0.08 & 0.28 & 0.66 & 0.34 \\
\midrule
MagicDrive \cite{gao2023magicdrive} & $224\times400$  & 12.92 & 28.36 & 21.95 & 51.46 & 17.10 & 5.25 & 0.57 & 1.14 & 1.95 & 1.22  &  0.10 & 0.25 & 0.70 & 0.35  \\
{\method} & $224\times400$ & 16.63 & 30.57 & 26.16 & 58.98 & 20.22 & 8.83 & 0.55 & 1.08 & 1.85 & 1.16  & 0.08 & 0.27 & 0.81 & 0.39  \\ 
{\method} & $336\times600$ & 24.11 & 37.27 & 29.92 & 66.20 & 23.78 & 12.76 & 0.53 & 1.05 & 1.79 & 1.12  & 0.03 & 0.20 & 0.65 & 0.29  \\
{\method} & $448\times800$ & \textbf{26.00} & \textbf{38.66} & \textbf{30.98} & \textbf{67.76} & \textbf{24.87} & \textbf{13.46} & \textbf{0.52} & \textbf{1.02} & \textbf{1.72} & \textbf{1.09}  & \textbf{0.02} & \textbf{0.17} & \textbf{0.55} & \textbf{0.25}  \\ \bottomrule
\end{tabular}
}
\caption{Comparison of generation fidelity. The data synthesis conditions are from the nuScenes validation set. All results are computed by using the official implementation and checkpoints of UniAD.}
\label{tab:uniad_val}
\end{center}
\end{table*}

\subsection{More Related Works}
\subsubsection{ Diffusion-based conditional generation}
Diffusion models have revolutionized generative tasks in both the image and video domains~\cite{podell2023sdxl,saharia2022photorealistic,li2023videogen,zheng2024open,he2024id}. In the realm of image generation, models like Stable Diffusion~\cite{rombach2022high}, PixArt~\cite{chen2023pixart}, and Flux~\cite{flux2024} are capable of generating high-quality images, whereas video diffusion techniques, exemplified by Stable Video Diffusion~\cite{blattmann2023stable}, OpenSora~\cite{zheng2024open} and Cogvideox~\cite{yang2024cogvideox}, mitigate challenges associated with temporal consistency and motion dynamics. While traditional Text-to-Image (T2I) and Text-to-Video (T2V) methods~\cite{ramesh2022hierarchical,rombach2022high,chen2023pixart,guo2023animatediff,blattmann2023align,yang2024cogvideox,zheng2024open} often struggle to provide precise control over the generated content, ControlNet~\cite{zhang2023adding} merges and addresses this limitation by training a control network that copies parts of the pre-trained main model to introduce control conditions, such as edge maps, segmentation masks, and poses.

In the field of autonomous driving, precise control over video generation plays a vital role in developing realistic simulations. Recently, diffusion-based controllable methods such as MagicDrive~\cite{gao2023magicdrive}, DrivingDiffusion\cite{li2023drivingdiffusion}, and Panacea~\cite{wen2024panacea} have emerged for street-view scene generation. These approaches integrate 3D bounding boxes, Bird’s Eye View (BEV) maps, ego trajectories, and camera poses to synthesize multi-view street scenes. 
To take advantage of the strong spatiotemporal modeling of transformers \cite{chen2023pixart, zheng2024open}, MagicDriveDiT~\cite{gao2024magicdrivedit} integrate the DiT architecture~\cite{peebles2023scalable, jiang2024dive} with 3D VAEs~\cite{zheng2024open,yang2024cogvideox} to manage spatiotemporal latent representations. 
Different from the above methods, we propose a motion-aware autoregressive architecture, which introduce perspective guidance and object-wise position encoding to improve controllability and motion-aware temporal attention to improve temporal coherence and seamless video generation. It can well adapt to various generative base models, highlighting its broad applicability to the autonomous driving community.

\subsection{Implementation Details} \label{Sec.imp}
\subsubsection{Post-processing strategy}
\label{Appendix 2.1}
We have experimentally observed that using overlapping frames within a sliding window can slightly enhance generation stability. Furthermore, integrating with DriveArena \cite{yang2024drivearena}, it is necessary to overlap a few frames to align the output frequency of different components (detailed below). Therefore, we also propose an optional post-processing strategy that utilizes overlapping frames within the sliding windows.
Specifically, at the $t$-step of the denoising process for the current video clip, we replace the noised latents $\textbf{Z}^t_T[\colon N]$ with latents from the previous video clip, denoted as $\sqrt{\overline{\alpha}_t}\cdot \widehat{\textbf{Z}}^{0}_T[-N\colon] + \sqrt{1-\overline{\alpha}_t}\cdot \epsilon_t$, before feeding them into the denoising U-Net. Here, $\widehat{\textbf{Z}}^{0}_T$ denotes the latents extracted using the VAE encoder, while $\epsilon_t$ represents the Gaussian noise at the $t$-step. $\overline{\alpha}_t$ is the hype-parameters in the diffusion process. In this way, we ensure that the first $N$ frames of the current video are as consistent as possible with the last $N$ frames of the previous clip for improved coherence.

\subsubsection{Training and Inference} We train the newly added modules on eight A100 GPUs using the AdamW optimizer \cite{loshchilov2017decoupled} with a learning rate of 8e-5. The training process consists of two stages. In the first stage, we train the single-frame version without the motion-aware temporal attention module for 100 epochs with a total batch size of 24. The training objective and hype-parameters are consistent with \cite{gao2023magicdrive}. In the second stage, we focus solely on training the temporal module for another 100 epochs, using a total batch size of 8. The motion frames are randomly sampled from the previous 5 frames with GT values. For higher-resolution models, we train for 50,000 iterations, initializing with the weights from the smaller-resolution model. For the variants employing DiT \cite{zheng2024open} and 3D VAE \cite{yang2024cogvideox} as the base models, we trained from scratch for 300K iterations at a resolution of $224\times400$, utilizing 8 A100 GPUs with a batch size of 4 per GPU. Subsequently, we proceeded to train at a larger resolution of $448 \times 800$ for an additional 50K iterations. Finally, we leveraged the pretrained weights to train the video version at $448 \times 800$ resolution for another 50K iterations. 

Following the approach outlined in MagicDrive \cite{gao2023magicdrive, gao2024magicdrivedit}, we employ the UniPC \cite{zhao2024unipc} scheduler for 20 steps with the base model of SD V1.5 and the rectified flow \cite{liu2022flow} scheduler for 30 steps with the base model of DiT, applying a CFG (Classifier-Free Guidance) scale of 2.0 to generate the multiview videos. The motion frames are sampled from previously generated video clips. When generating long videos, for the first video clip, we use the single-frame model to generate the initial frame as the motion frames. By default, the length of the overlapping frames in the post-processing strategy is set to 2. Note that we do not train a new model for different video lengths; all videos of varying lengths are generated using the model trained in short sequences.

\begin{figure}[t]
    \centering
    \includegraphics[width=\linewidth]{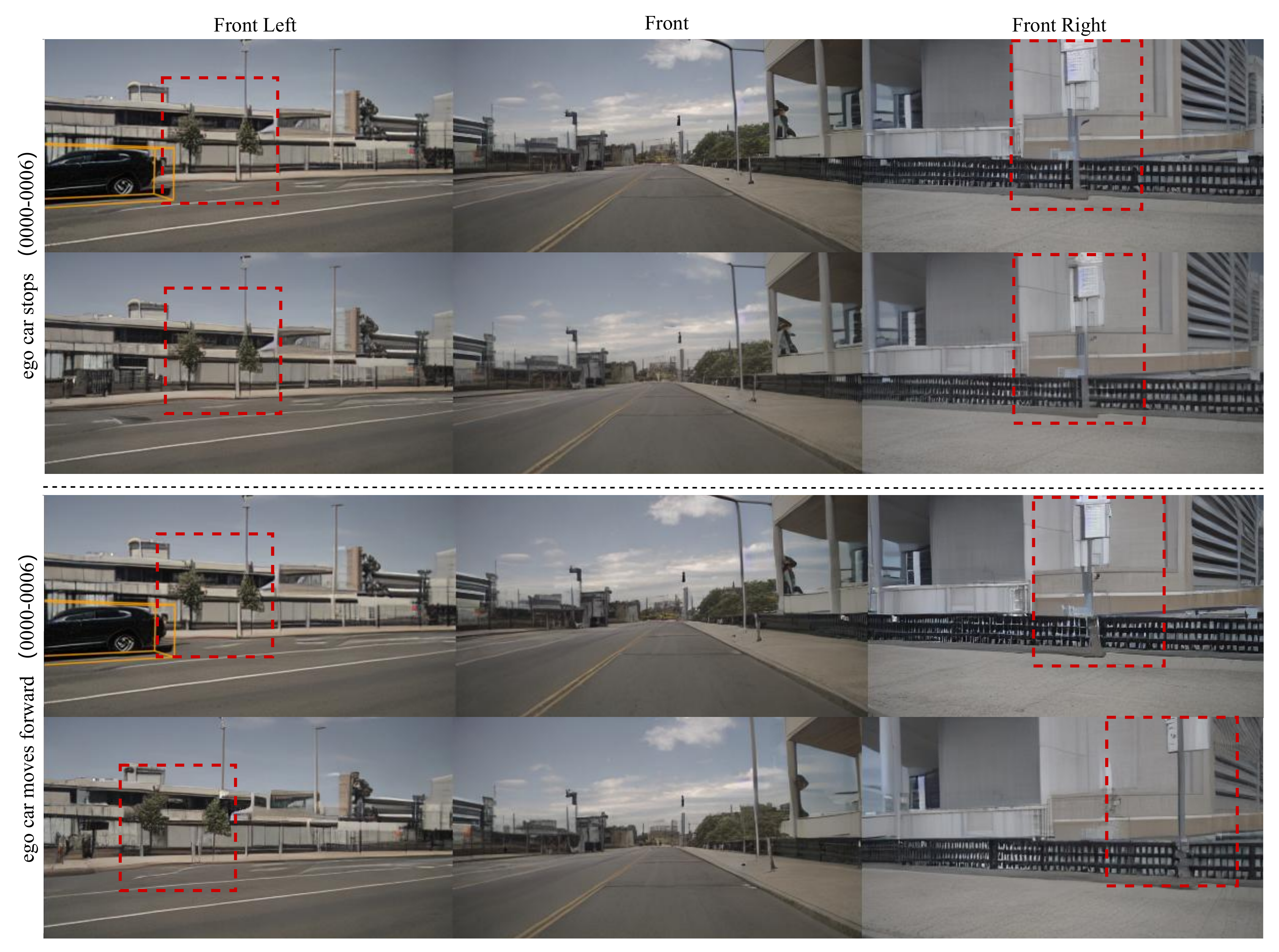}
    \caption{The ego pose can influence changes in the background, as illustrated by the red circles.}
    \label{fig:vis_2}
\end{figure}

\subsubsection{Integration with DriveArena.} DriveArena \cite{yang2024drivearena} offers a modular platform that can be integrated with any vision-based driving agent for both open-loop and closed-loop simulations. It comprises two key components: (1) Traffic Manager, which processes high-definition maps downloaded from the internet to create diverse urban layouts, manages vehicle movements and traffic flow, and handles collision detection. (2) World Dreamer, a generative model that generates photorealistic multi-view camera images corresponding to the simulation state and adjusts controllable parameters based on specified prompts. All these components exchange data through the network interface. The Traffic Manager runs at 10 Hz while the common vision-based agents such UniAD \cite{hu2023planning} and VAD \cite{jiang2023vad} take multiview images at 2 Hz. Therefore, it is necessary to synchronize the Traffic Manager, our 7-frame {\method}, and the driving agents. We utilize a queue of length 7 to cache the data from the Traffic Manager, which is sent to our {\method} at 2 Hz. In this manner, {\method} receives 7-frame data each time, with the previous 2 frames overlapping with those from the last iteration. Subsequently, the last frame is taken as the keyframe and sent to the driving agents for planning. In the open-loop mode, the Traffic Manager generates trajectory for ego vehicles. While in closed-loop mode, the trajectory outputted by the driving agents at 2 Hz, consisting of six path points over the next 3 seconds, is interpolated to create a 10 Hz trajectory, which is then directly used for ego vehicle control. Through a motion-aware autoregressive generation paradigm, our {\method} supports long-term multi-view generation. However, it is inevitable that accumulated errors gradually arise during the iterative process of the simulator. To alleviate the above issues, we empirically used motion frames sampled from the previous clip as conditions for the single-frame version, refining these frames to reduce potential accumulated errors.

\paragraph{Mask‐Shift Mechanism in DreamForge-DiT}  
We employ a hybrid masking approach at the feature level during training in DreamForge-DiT, randomly selecting one of two masking strategies for each training sample (50\% probability for each). 
\begin{itemize}
\item \textbf{Random Masking:} Following Open-Sora 1.2 \cite{zheng2024open}, a random subset of frame positions in the sequence are designated as targets (mask set to \texttt{True}), while the remaining frames serve as context (mask set to \texttt{False}). This strategy encourages the model to learn generation conditioned on diverse temporal contexts.

 \item \textbf{Autoregressive Masking:}  The first $N$ frames are fixed as context (setting them to \texttt{False} in the mask) ,and the model is tasked with generating the subsequent $T - N$ frames (setting them to \texttt{True}), enforcing sequential, autoregressive generation from past to future.
\end{itemize}
By alternating between these two strategies during training, we found that DreamForge-DiT demonstrates superior performance in long video generation, producing sequences that effectively maintain both local detail and global coherence across extended autoregressive temporal horizons.

\begin{figure*}[tbp]
    \centering
    \includegraphics[width=\linewidth]{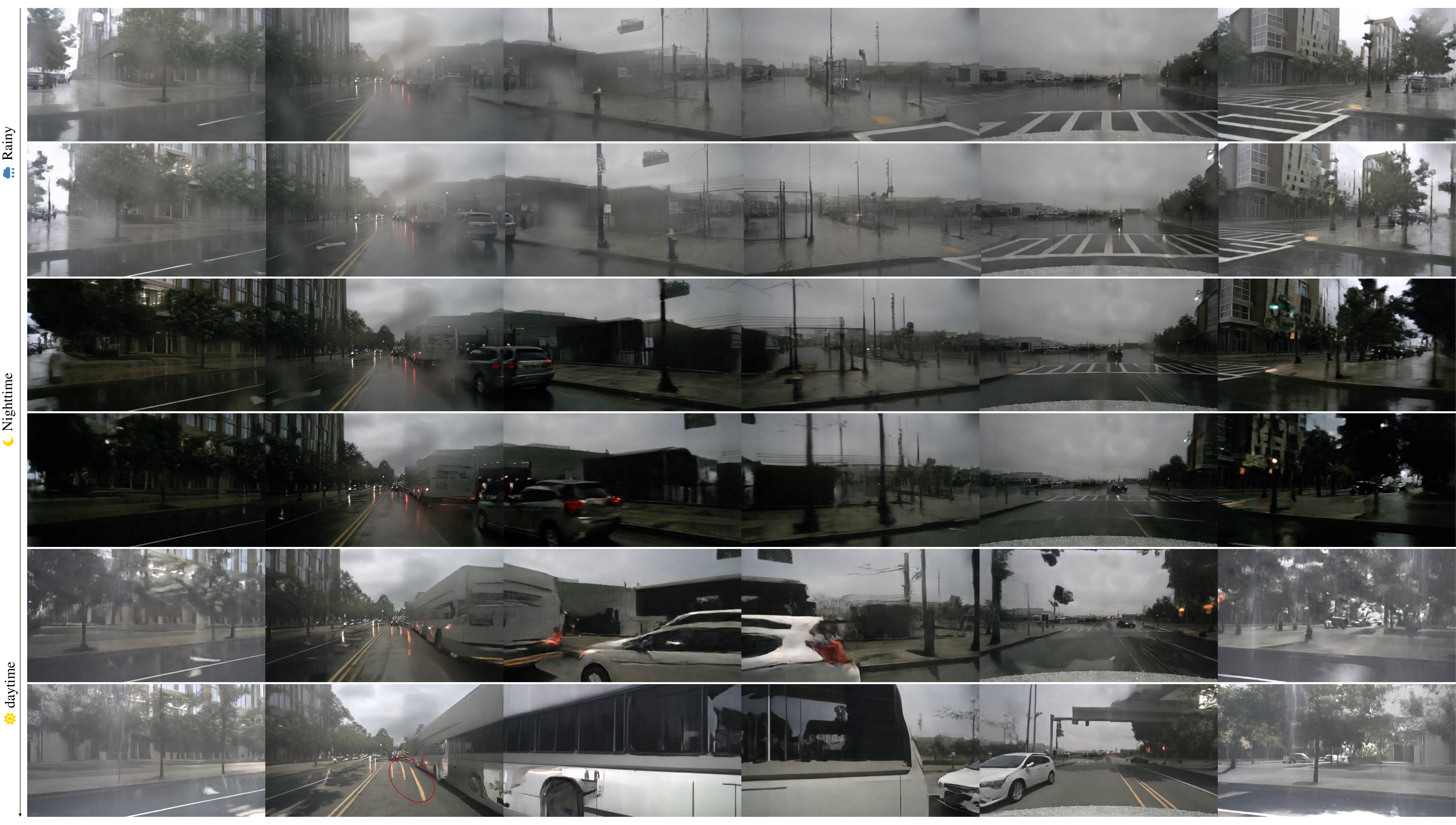}
    \caption{By combining autoregressive generation and text prompts for different weather conditions, our model can produce videos that showcase a seamless weather transition within continuous clips. We also find the model still has some limitations such as the accurate transition of light reflection as shown in the red circle, which will be a focus of our future improvement. }
    \label{fig:vis_shift}
    \vspace{-15pt}
\end{figure*}

\subsection{Dataset and Metrics} \label{Sec.data}
\subsubsection{Dataset.} 
We utilize the nuScenes dataset \cite{caesar2020nuscenes} to train our controllable multiview street view video generation model {\method}. The nuScenes dataset provides 6 camera views at 12 Hz, offering a 360-degree perspective of the scenes. It includes 750 scenes for training and 150 scenes for validation, encompassing different cities and a variety of lighting and weather conditions, such as daytime, nighttime, sunny, cloudy, and rainy scenarios. Since the nuScenes dataset only provides annotations at 2 Hz, we employ ASAP \cite{wang2023we} to generate interpolated annotations at 12 Hz. Additionally, we annotated each scene using GPT-4, providing detailed descriptions that include elements like time, weather, street style, road structure, and appearance. These descriptions serve as conditions for text input.
\subsubsection{Metircs.} 
We use FID \cite{heusel2017gans} and FVD \cite{unterthiner2018towards} to assess the quality of the generated images and videos. Additionally, we evaluate the sim-to-real gap by measuring performance on the generated scenes in downstream tasks, including 3D object detection (mAP and NDS), BEV segmentation (mIoU), and end-to-end planning (L2 and Collation rate). Following \cite{Dauner2024navsim, carla2023leaderboard, yang2024drivearena}, we employ the PDM Score (PDMS) and Arena Driving Score (ADS) to evaluate the performance of driving agents in both open-loop and closed-loop modes within the DriveArena simulator \cite{yang2024drivearena}. PDMS~\citep{Dauner2024navsim} assesses the trajectory output at each timestep, incorporating penalties for driving without collisions (NC) with road users and compliance with the drivable area (DAC). It also includes a weighted average of factors such as ego progress (EP), time-to-collision (TTC), and comfort (C). Building on PDMS, ADS \cite{yang2024drivearena} is calculated by multiplying the modified PDMS by the route completion score \cite{carla2023leaderboard}.

\subsection{More Results} \label{Sec:results}
\subsubsection{Quantitative Results.} 
We also utilize UniAD \cite{hu2023planning} as an evaluator for the generated scenes to compare various metrics, including 3D object detection, BEV map segmentation, and planning, as illustrated in Table \ref{tab:uniad_val}. We can see that, compared to the state-of-the-art method MagicDrive, our method outperforms it in nearly all metrics, except for a slight lag in the collision rate. Interestingly, we found that when we increase the input resolution, the collision rate in the scenes generated by our method is lower than that observed in the real data. Furthermore, the L2 error is also smaller when using the same input resolution of $224 \times 400$.

\begin{table*}[ht]
    \centering
    \begin{tabular}{l l c c c c c c}
        \toprule
        Scenario & Base Model & NC $\uparrow$ & DAC $\uparrow$ & EP $\uparrow$ & TTC $\uparrow$ & C $\uparrow$ & PDMS $\uparrow$ \\
        \midrule
        DriveArena & SD 1.5 & 0.792 & 0.942 & 0.738 & 0.771 & \textbf{0.749} & 0.636 \\
        DreamForge & SD 1.5 & 0.843 & \textbf{0.958} & 0.728 & 0.829 & 0.704 & 0.669 \\
        DreamForge & DiT, 3D VAE & \textbf{0.8721} & 0.9220 & \textbf{0.7815} & \textbf{0.829} & 0.693 & \textbf{0.6753} \\
        \bottomrule
    \end{tabular}
    \caption{Open-loop evaluation results. The \textbf{\textit{bold}} means the best result.}
    \label{tab:open_loop}
\end{table*}

\begin{table}[t]
    \centering
    \resizebox{\linewidth}{!}{%
    \begin{tabular}{c|c|cc}
    \toprule
    \textbf{Scenario} & \textbf{Method} & \textbf{PDMS} $\uparrow$ & \textbf{ADS} $\uparrow$ \\ 
    \midrule
    \multirow{3}{*}{\parbox{2cm}{\texttt{singapore-}\\\texttt{onenorth}}} 
    & DriveArena & 0.7615 & 0.1282 \\
    & DreamForge & 0.8102 & 0.1197 \\  
    & DreamForge-DiT & 0.8071 & 0.1207 \\  
    \midrule
    \multirow{3}{*}{\parbox{2cm}{\texttt{boston-}\\\texttt{seaport}}}  
    & DriveArena & 0.4952 & 0.0450 \\  
    & DreamForge & 0.7401 (\textbf{+ 0.2449}) & 0.0760 (\textbf{+ 0.0310}) \\
    & DreamForge-DiT & 0.8192 (\textbf{+ 0.3240}) & 0.1448 (\textbf{+ 0.0998}) \\
    \bottomrule
    \end{tabular}}
    \caption{Close-loop evaluation results with UniAD in DriveArena}
    \label{tab:close_loop}
    \vspace{-4pt}
\end{table}

\subsubsection{Visualizations.} 
\label{Appendix 4.2}
\begin{figure*}[tbp]
    \centering
    \includegraphics[width=0.98\linewidth]{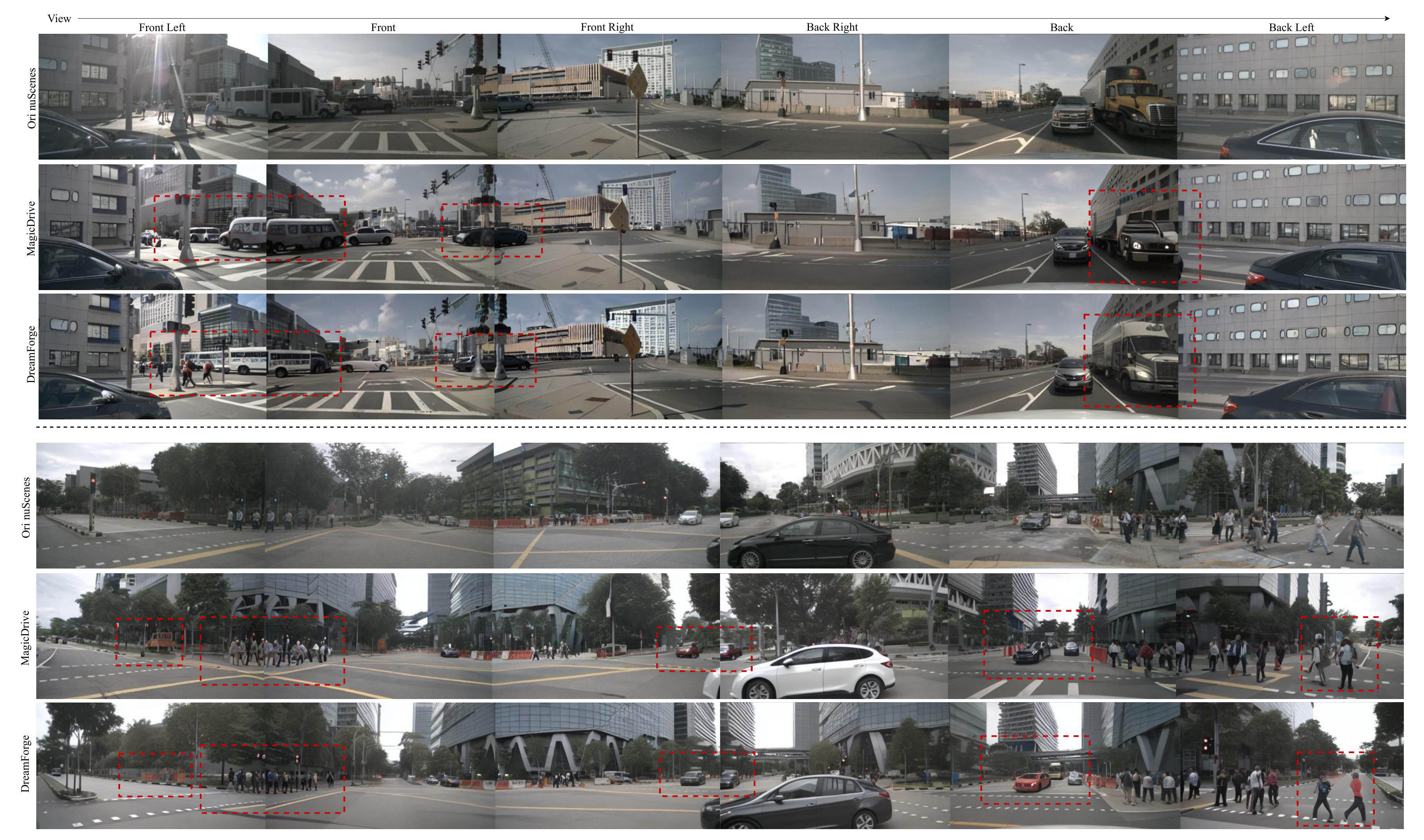}
    \caption{Visual comparison of foreground generation. The illustrations demonstrate that our {\method} outperforms the baseline in object generation. Additionally, the object consistency across different views is also improved with our method.}
    \label{fig:vis_0}
    \vspace{-15pt}
\end{figure*}

\begin{figure*}[tbp]
    \centering
    \includegraphics[width=0.98\linewidth]{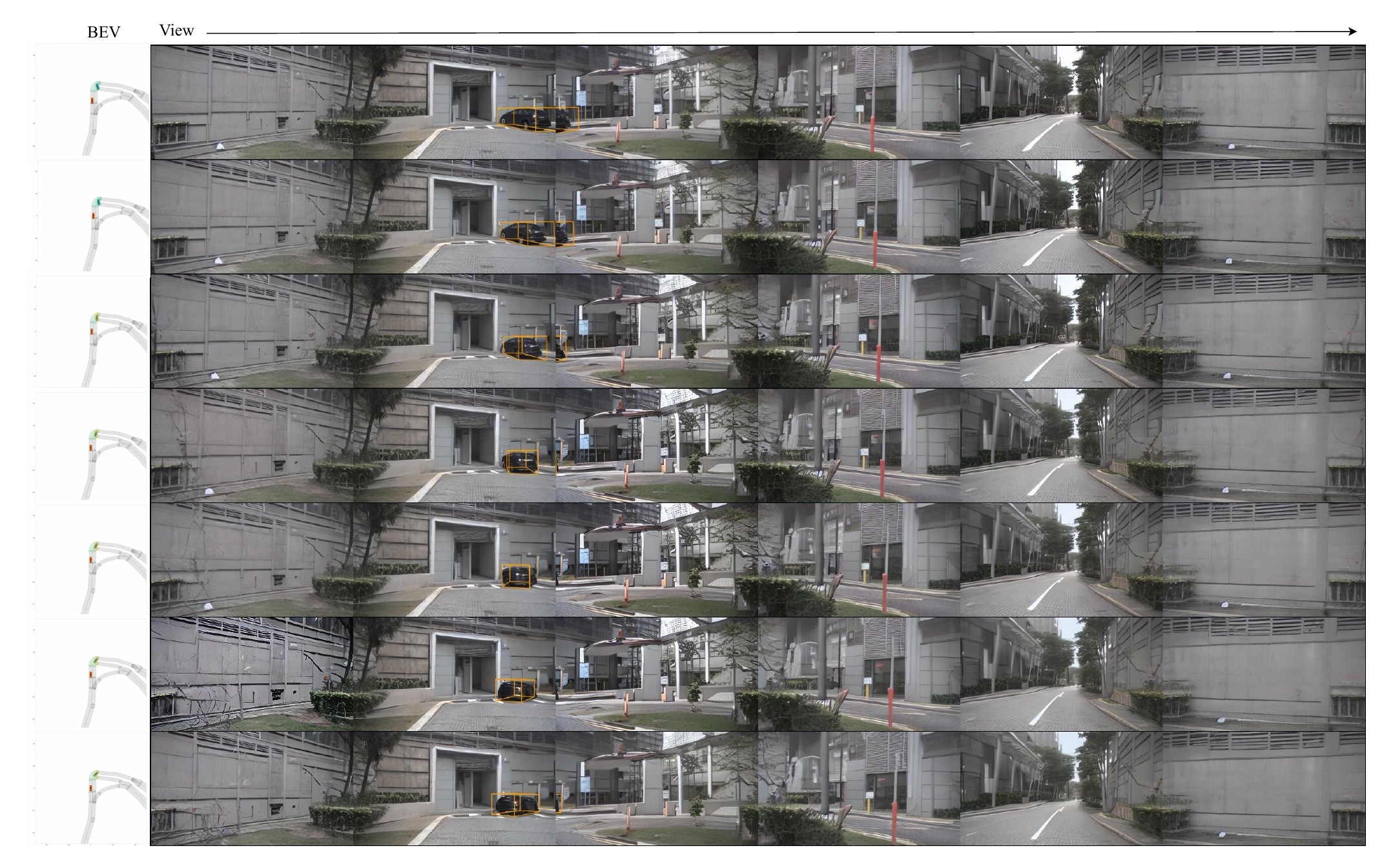}
    \caption{The visualizations illustrate that our method can adapt to the road layouts and 3D bounding boxes generated by DriveArena \cite{yang2024drivearena}.}
    \label{fig:vis_bev}
\end{figure*}

\begin{figure*}[tbp]
    \centering
    \includegraphics[width=0.98\linewidth]{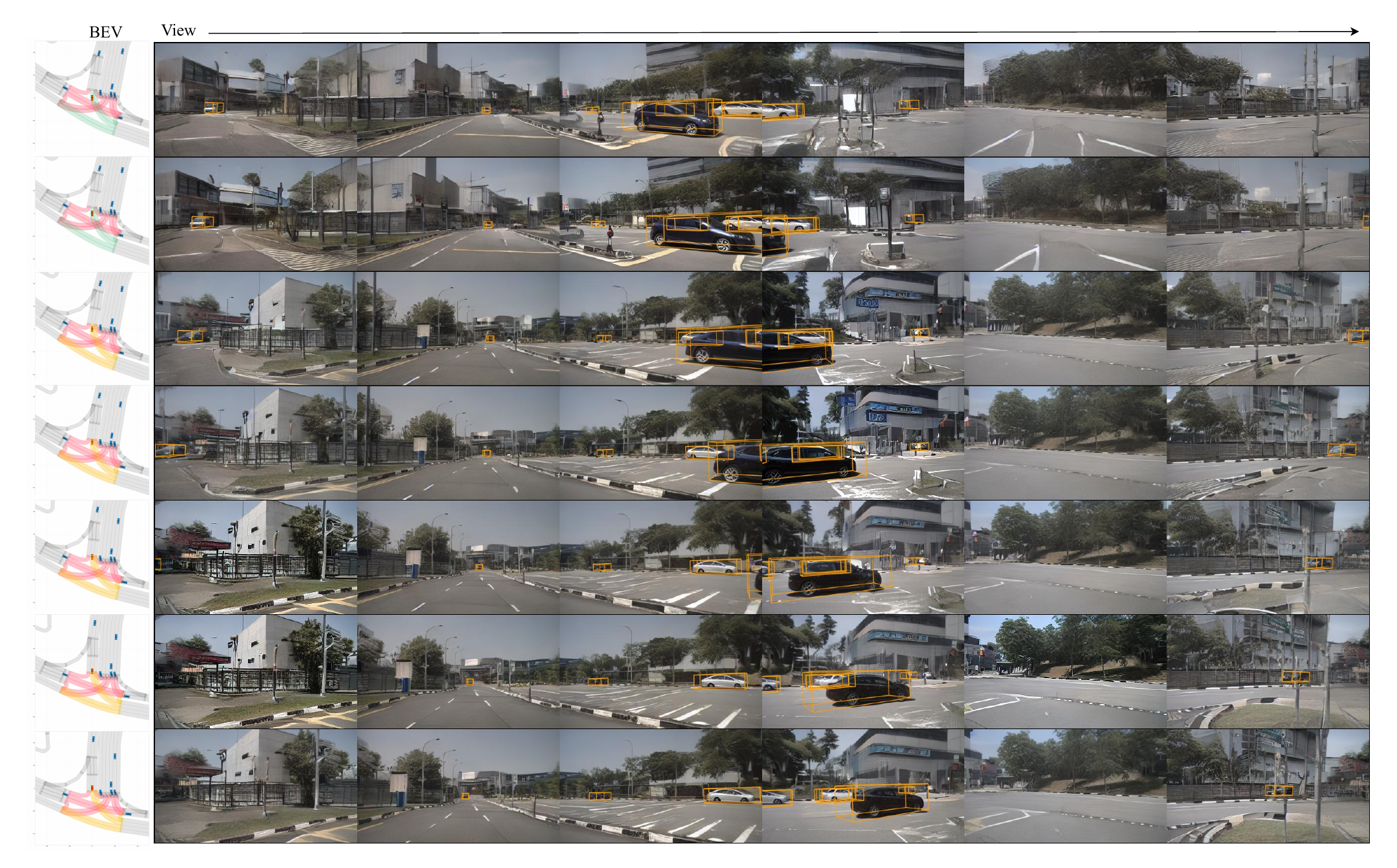}
    \caption{Our {\method} can adapt to the complex road layouts and 3D bounding boxes generated by DriveArena \cite{yang2024drivearena}.}
    \label{fig:vis_bev_2}
    \vspace{-15pt}
\end{figure*}

\begin{figure*}[tbp]
    \centering
    \includegraphics[width=0.98\linewidth]{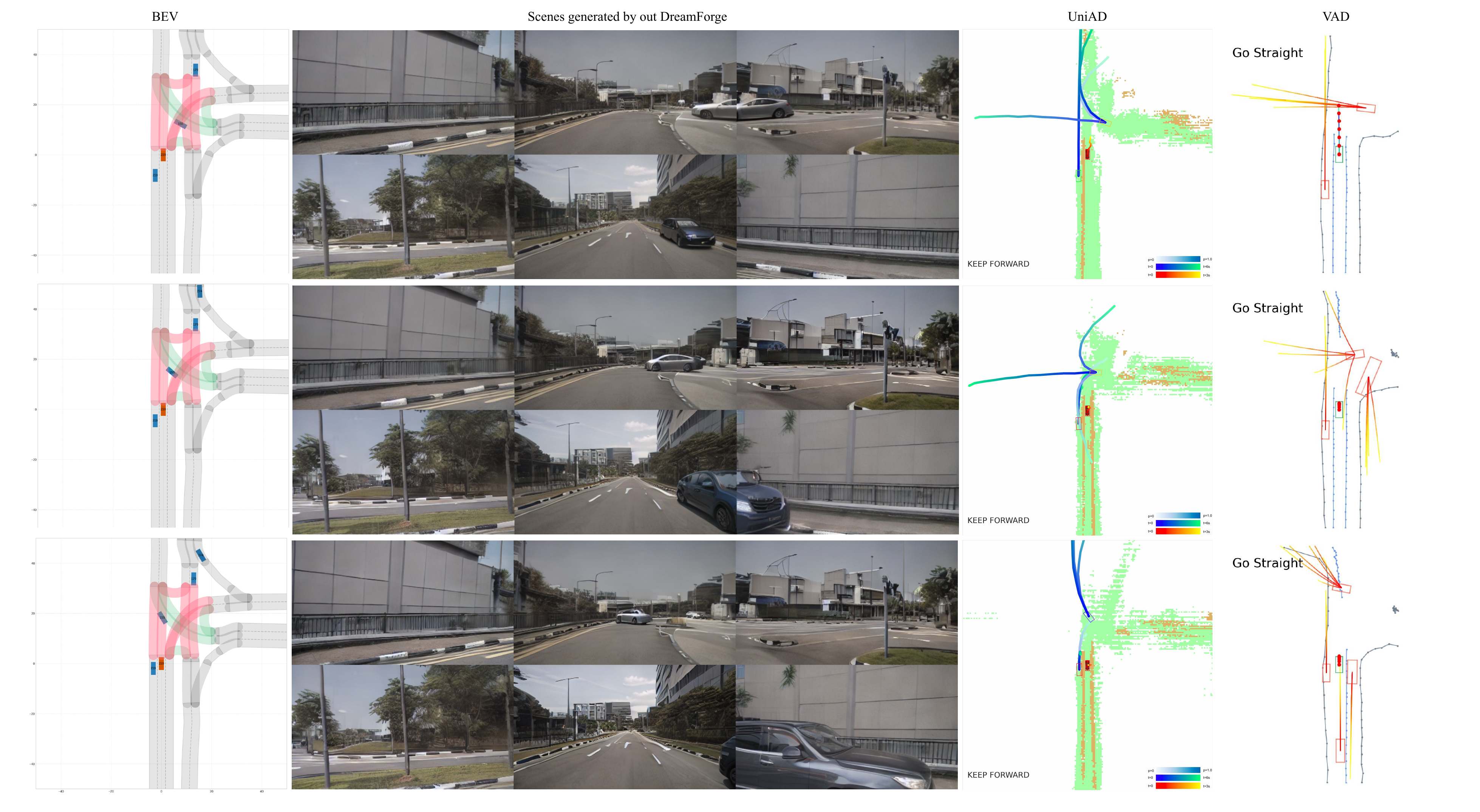}
    \caption{Visualizations of the simulation within DriveArena \cite{yang2024drivearena}. From left to right: BEV layouts from DriveArena; Keyframes generated by our {\method}; BEV predictions of UniAD \cite{hu2023planning} and VAD \cite{jiang2023vad}.}
    \label{fig:vis_agent}
    \vspace{-10pt}
\end{figure*}

\begin{figure*}[tbp]
    \centering
    \includegraphics[width=\linewidth]{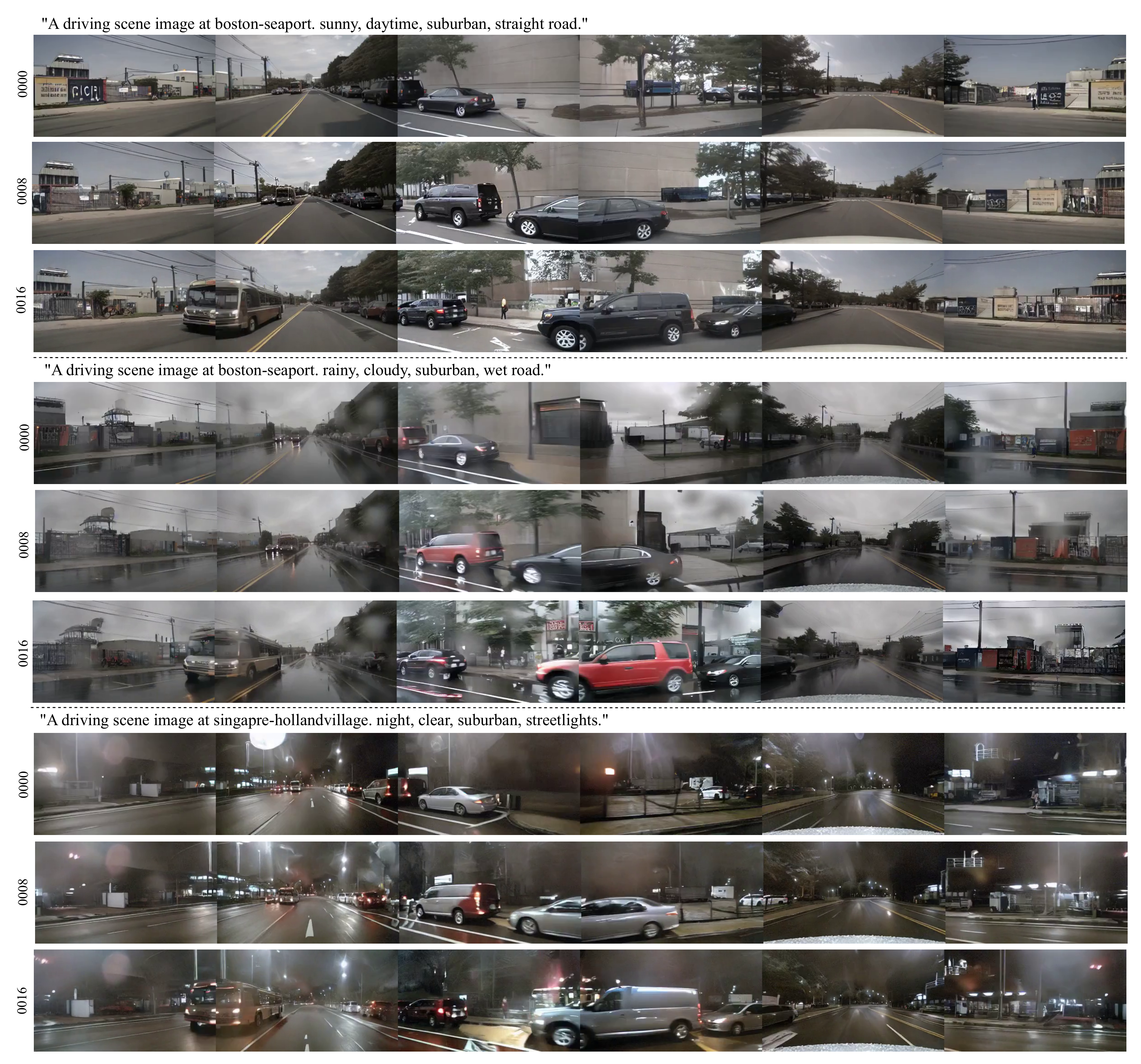}
    \caption{Visualizations featuring various text prompts for different weather conditions, such as sunny, rainy, and night. For a better view, we visualize several sampled keyframes from the generated videos.}
    \label{fig:vis_text}
    \vspace{-15pt}
\end{figure*}

\begin{figure*}[tbp]
    \centering
    \includegraphics[width=0.98\linewidth]{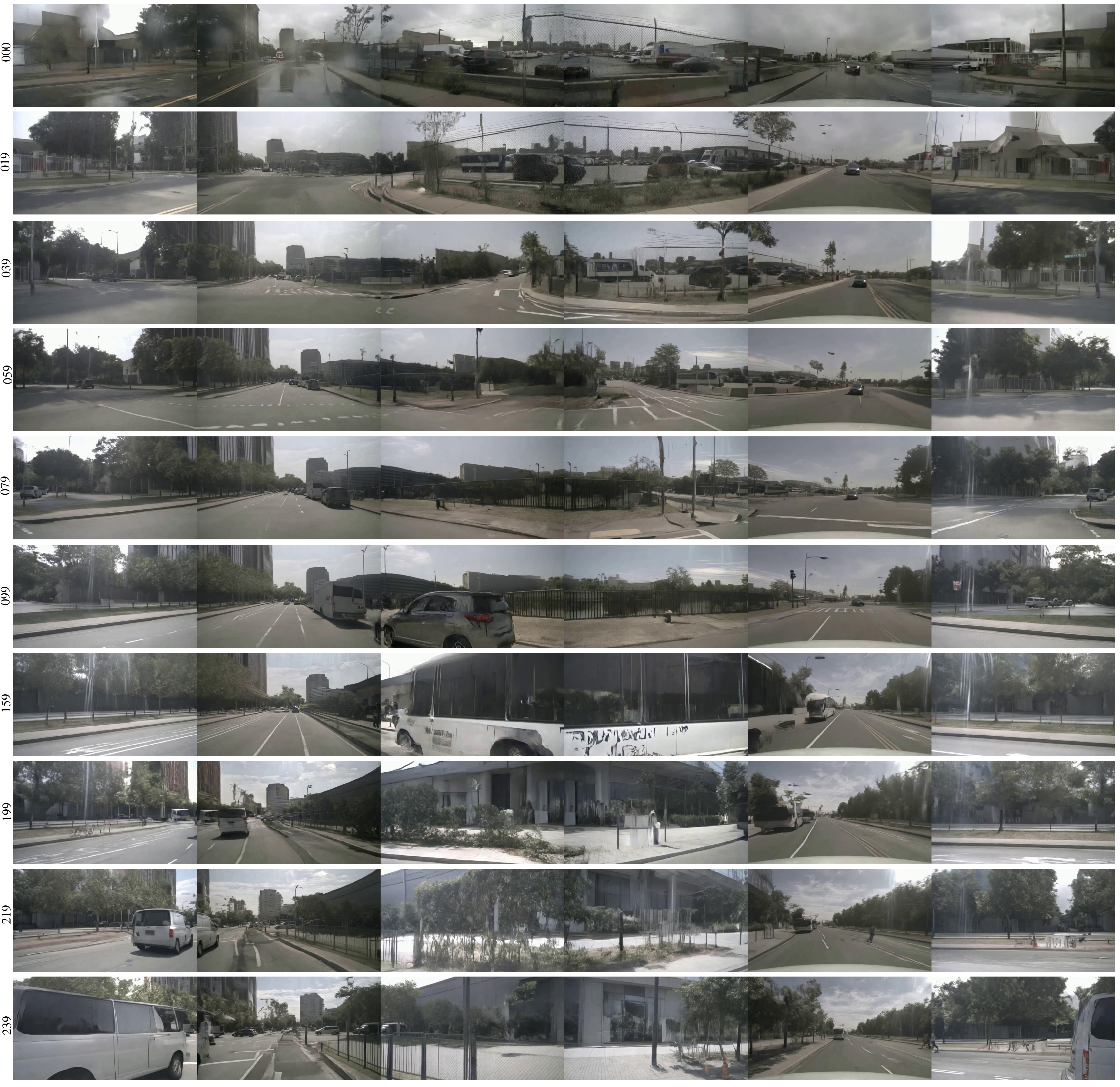}
    \caption{The visualizations illustrate that our model can produce video sequences that are both multiview and temporally coherent, generating realistic content in an autoregressive manner.}
    \label{fig:vis_long}
\end{figure*}

\begin{figure*}[tbp]
    \centering
    \includegraphics[width=0.98\linewidth]{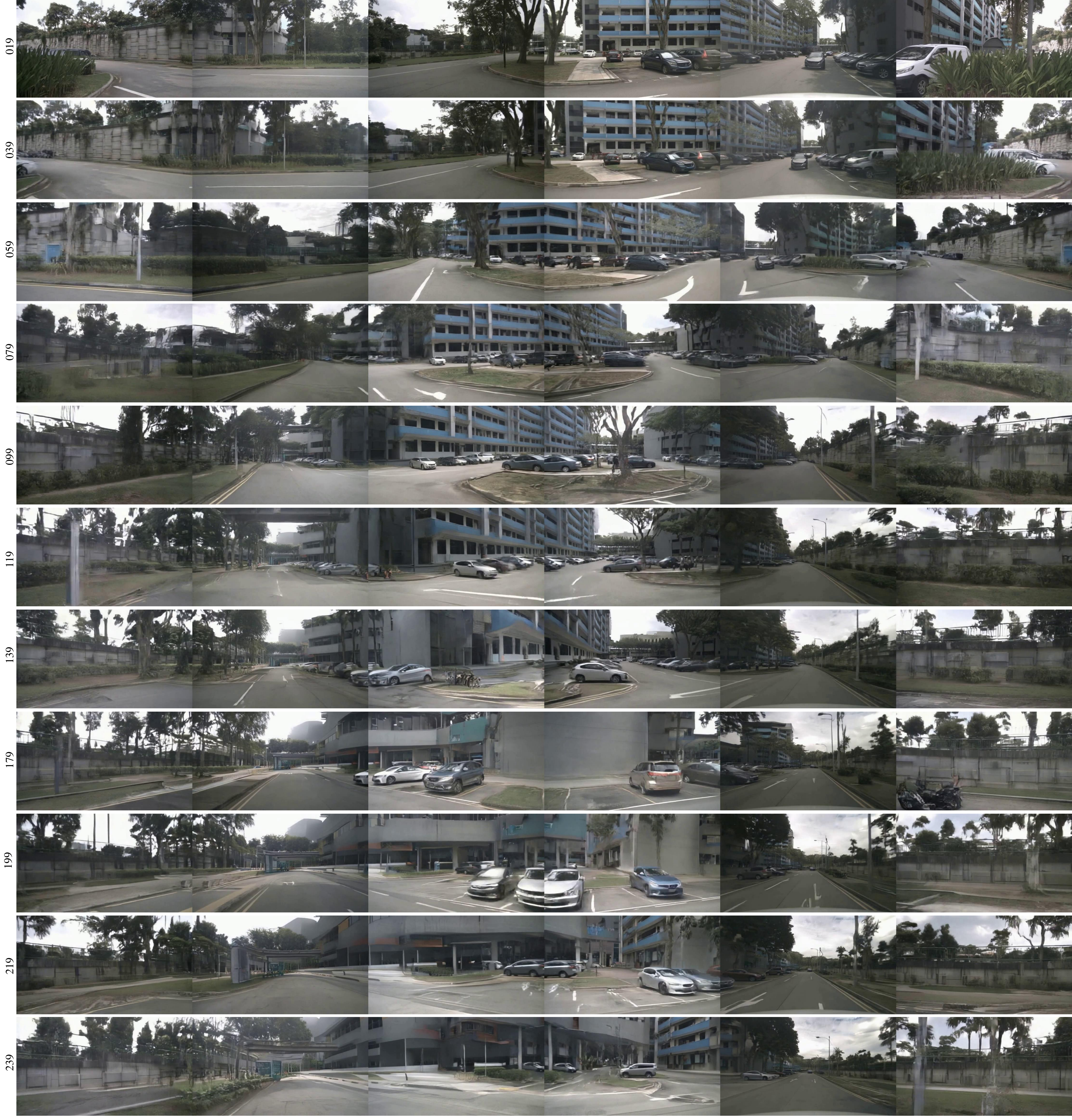}
    \caption{The visualizations illustrate that our model can produce video sequences that are both multiview and temporally coherent, generating realistic content in an autoregressive manner.}
    \label{fig:vis_long_2}
\end{figure*}

\textbf{3D controllability}. We provide more cases in Fig.~\ref{fig:vis_0} to demonstrate that our {\method} performs better in accurate object generation and maintaining consistency in street views. The visualizations illustrate that our method produces buses, trucks, and people with improved appearance, particularly in the cross-view area. We also offer visualizations that utilize the road layouts and 3D bounding boxes generated by DriveArena~\cite{yang2024drivearena}. The results, presented in Fig.~\ref{fig:vis_bev} and Fig.~\ref{fig:vis_bev_2}, demonstrate that our method can generate controllable urban scenes by modifying the layout of the roads and the box boundaries of objects. We further explore the effect of ego poses on background changes. As shown in Fig.~\ref{fig:vis_2}, by modifying the ego car's pose (from ``stop" to ``move forward"), we observe that the generated background exhibits significant changes, demonstrating that the network can effectively extract motion cues from the ego poses. \\
\textbf{Weather alteration}. The examples in Fig.~\ref{fig:vis_text} demonstrate how {\method} transforms scenes to replicate diverse weather conditions and times of day, highlighting its controllability in modifying scene descriptions. Additionally, Fig.~\ref{fig:vis_shift} showcases a seamless weather transition within continuous video clips. These cases clearly illustrate that the proposed motion-aware autoregressive generation method provides a highly flexible approach to video appearance control. It not only supports the generation of varying weather conditions for the same scene but also facilitates smooth and natural appearance transitions within continuous video sequences. We also find that the model has some limitations, such as its inability to effectively handle changes in light reflection. For instance, as shown in the red circle in Fig.~\ref{fig:vis_shift} it fails to accurately represent the transition of light reflection on lane lines from nighttime to daytime. Addressing this issue will be our future work.\\
\textbf{Long video generation.} We present additional case studies in Fig.~\ref{fig:vis_long} and Fig.~\ref{fig:vis_long_2} that highlight the 3D controllability of our model, particularly its ability to maintain high video fidelity and temporal consistency across generated frames. These examples illustrate that our model can produce video sequences that are both multiview and temporally coherent, generating realistic content in an autoregressive manner. This capability demonstrates its potential applications in realistic autonomous driving simulation, where accurate and consistent video generation is crucial for the training and testing of autonomous systems. \\
\textbf{Simulation within DriveArena}. We illustrate the visualizations of the simulations within DriveArena. Our {\method} receives sequential data from the Traffic Manager in DriveArena and sends generated scenes of keyframes to the driving agents at 2 Hz, as stated in the implementation details. Without loss of generality, we present cases where the driving agents UniAD \cite{hu2023planning} and VAD \cite{jiang2023vad} operate in an open-loop mode in Fig.~\ref{fig:vis_agent}. The results indicate that predictions from the driving agents regarding the road network and vehicle tracking are fundamentally accurate, with UniAD demonstrating superior perception results on the generated scenes. Additionally, we integrated with the DriveArena \cite{yang2024drivearena} simulation platform to demonstrate the practical benefits of DreamForge-DiT. In both open-loop and closed-loop evaluations as shown in Table \ref{tab:open_loop} and \ref{tab:close_loop}, the model provided more reliable and realistic scenarios for testing vision-based driving agents. The enhanced temporal coherence and controllability of DreamForge-DiT resulted in more accurate simulations of real-world driving conditions.

\end{document}